%% file: abe-arxiv.tex
\documentclass[runningheads]{llncs}
\usepackage{graphicx}
\usepackage{amsmath,amssymb} % define this before the line numbering.
\usepackage{color}
\usepackage{cellspace} %
\usepackage{subfigure}
\usepackage[sort]{cite}
\usepackage[symbol]{footmisc}

\begin{document}
% \linenumbers
%\pagestyle{headings}
%\mainmatter
%\def\ECCV18SubNumber{1314}  % Insert your submission number here

\title{Attention-based Ensemble for \\Deep Metric Learning}

\titlerunning{Attention-based Ensemble for Deep Metric Learning}

\authorrunning{W. Kim, B. Goyal, K. Chawla, J. Lee, K. Kwon}

\author{Wonsik Kim, Bhavya Goyal, Kunal Chawla, Jungmin Lee, Keunjoo Kwon}
\institute{Samsung Research, \\
Samsung Electronics\\
\email{ \{wonsik16.kim, bhavya.goyal, kunal.chawla, jm411.lee, keunjoo.kwon\}@samsung.com}
}

\maketitle

% \DeclareRobustCommand\onedot{\futurelet\@let@token\@onedot}
% \def\@onedot{\ifx\@let@token.\else.\null\fi\xspace}
\def\onedot{.}
\def\eg{\emph{e.g}\onedot} \def\Eg{\emph{E.g}\onedot}
\def\ie{\emph{i.e}\onedot} \def\Ie{\emph{I.e}\onedot}
\def\cf{\emph{c.f}\onedot} \def\Cf{\emph{C.f}\onedot}
\def\etc{\emph{etc}\onedot} \def\vs{\emph{vs}\onedot}
\def\wrt{w.r.t\onedot} \def\dof{d.o.f\onedot}
\def\etal{\emph{et al}\onedot}
\def\ensemblefunc{b}
\def\Ensemblefunc{B}
\def\ensemblesub{m}
\def\Ensemblesub{M}

\input{abstract.tex}

\input{introduction.tex}

\input{related-works.tex}

\input{proposed-method.tex}

\input{implementation.tex}

\input{evaluation.tex}

\input{experiments.tex}

\input{conclusion.tex}

\bibliographystyle{splncs04}
\bibliography{metric}

\clearpage
\appendix
\setcounter{secnumdepth}{0}
\input{appendix.tex}

\end{document}

%% file: abstract.tex
\begin{abstract}
\vspace{-8mm}
Deep metric learning aims to learn an embedding function, modeled as deep neural network.
This embedding function usually puts semantically similar images close
while dissimilar images far from each other in the learned embedding space.
Recently, ensemble has been applied to deep metric learning to yield state-of-the-art results.
As one important aspect of ensemble, the learners should
be diverse in their feature embeddings. To this end, we propose an attention-based ensemble,
which uses multiple attention masks, so that each learner can attend to different parts of the
object. We also propose a divergence loss, which encourages diversity among the learners.
The proposed method is applied to the standard benchmarks of deep metric learning and
experimental results show that it outperforms the state-of-the-art methods by a significant
margin on image retrieval tasks.
\vspace{-4mm}
\keywords{attention, ensemble, deep metric learning}
\vspace{-4mm}
\end{abstract}

%% file: introduction.tex
\vspace{-5mm}
\section{Introduction}
\vspace{-2mm}
Deep metric learning has been actively researched recently.
In deep metric learning, feature embedding function is modeled as a deep neural network.
This feature embedding function embeds input images into feature embedding space with a certain desired condition.
In this condition, the feature embeddings of similar images are required to be close to each other while those of dissimilar images are required to be far from each other.
To satisfy this condition, many loss functions based on the distances between embeddings have been proposed
\cite{bell2015learning,hadsell2006dimensionality,chopra2005learning,weinberger2009distance,schroff2015facenet,oh2016deep,ustinova2016learning,sohn2016improved,song2017deep,law2017deep}.
Deep metric learning has been successfully applied in image retrieval task on popular benchmarks such as CARS-196~\cite{KrauseStarkDengFei-Fei_3DRR2013}, CUB-200-2011~\cite{WahCUB_200_2011}, Stanford online products~\cite{oh2016deep}, and in-shop clothes retrieval~\cite{liu2016deepfashion} datasets.

Ensemble is a widely used technique of training multiple learners to get a combined model, which performs better than individual models.
For deep metric learning, ensemble concatenates the feature embeddings learned by multiple learners which often leads to better embedding space
under given constraints on the distances between image pairs.
The keys to success in ensemble are high performance of individual learners as well as diversity among learners.
To achieve this objective, different methods have been proposed \cite{yuan2016hard,opitz2017bier}.
However, there has not been much research on optimal architecture to yield diversity of feature embeddings in deep metric learning.

\begin{figure}[t]
\begin{center}
    \mbox{%
    \subfigure[$3$-heads ensemble]{ \label{fig:conventional_ensemble} \includegraphics[width=0.45\linewidth]{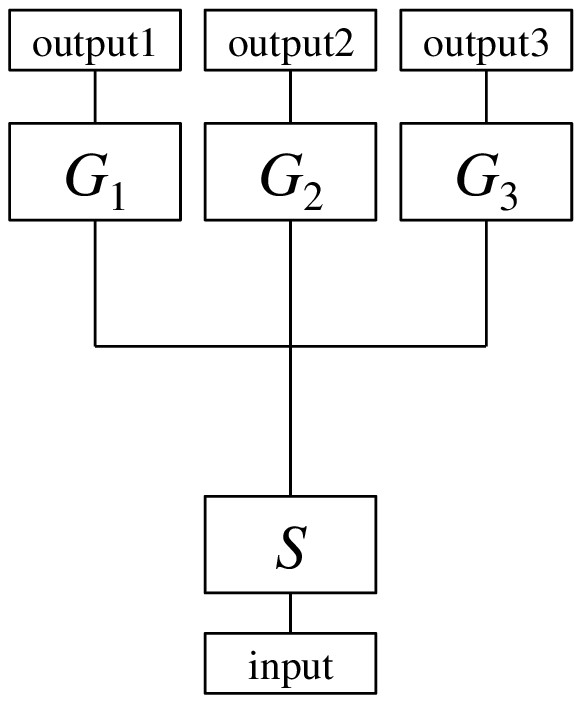}}
%    }
 %   \mbox{%
    \subfigure[Attention-based ensemble (ABE-$3$)]{ \label{fig:attentionbased_ensemble} \includegraphics[width=0.45\linewidth]{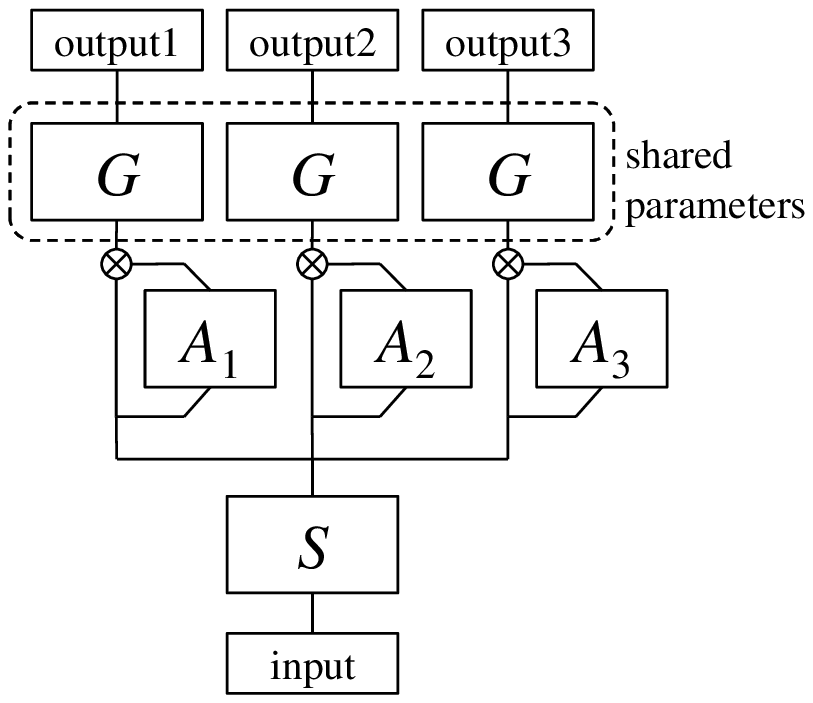}}
    }
\end{center}
\vspace{-7mm}
\caption{Difference between $M$-heads ensemble and attention-based ensemble.
Both assume shared parameters for bottom layers ($S$).
(a) In $M$-heads ensemble, different feature embedding functions are trained for different learners ($G_1, G_2, G_3$).
(b) In attention-based ensemble, single feature embedding function ($G$) is trained while each learner learns different attention modules ($A_1, A_2, A_3$)
\vspace{-6mm}
}
\label{fig:m_heads_and_m_way}
\end{figure}

Our contribution is to propose a novel framework to encourage diversity in feature embeddings.
To this end, we design an architecture which has multiple attention modules for multiple learners.
By attending to different locations for different learners, diverse feature embedding functions are trained.
They are regularized with divergence loss which aims to differentiate the feature embeddings from different learners.
Equipped with it, we present $M$-way attention-based ensemble (ABE-$M$) which learns feature embedding with $M$ diverse attention masks.
The proposed architecture is represented in Fig. \ref{fig:attentionbased_ensemble}.
We compare our model to our $M$-heads ensemble baseline~\cite{lee2015m}, in which different feature embedding functions are trained for different learners (Fig. \ref{fig:conventional_ensemble}),
and experimentally demonstrate that the proposed ABE-$M$ shows significantly better results with less number of parameters.

%% file: related-works.tex
\vspace{-2mm}
\section{Related works}
\vspace{-3mm}
\subsubsection{Deep metric learning and ensemble}
\label{sec:ensemble}

The aim of the deep metric learning is to find an embedding function $f: \mathcal{X} \rightarrow \mathcal{Y}$ which maps samples ${x}$ from a data space $\mathcal{X}$ to a feature embedding space $\mathcal{Y}$ so that $f({x}_i)$ and $f({x}_j)$ are closer in some metric when ${x}_i$ and ${x}_j$ are semantically similar.
To achieve this goal, in deep metric learning, contrastive \cite{hadsell2006dimensionality,chopra2005learning} and triplet \cite{weinberger2009distance,schroff2015facenet} losses are proposed.
%In this work, we are focusing on enhancing performance of \emph{deep metric learning} via ensemble strategy with attention mechanism.
Recently, more advanced losses are introduced such as lifted structured loss \cite{oh2016deep}, histogram loss \cite{ustinova2016learning}, N-pair loss \cite{sohn2016improved}, and clustering loss \cite{song2017deep,law2017deep}.
%Basically, those losses are designed to reduce distances between similar examples and increase distances between dissimilar examples.
%For readers who are interested in a detailed review of metric learning, we refer the survey work of Bellet \etal~\cite{bellet2013survey}.

Recently, there has been research in networks incorporated with ensemble technique, which report better performances than those of single networks.
Earlier deep learning approaches are based on direct averaging of the same networks with different initializations \cite{lee2015deeply,russakovsky2015imagenet} or training with different subsets of training samples \cite{Szegedy_2015_CVPR,sutskever2014sequence}.
%\cite{krizhevsky2012imagenet,russakovsky2015imagenet,lee2015deeply}
Following these former works, \emph{parameter sharing} is introduced by Bachman \etal~\cite{bachman2014learning} which is called \emph{pseudo-ensembles}.
Another \emph{parameter sharing} ensemble approach is proposed by Lee \etal~\cite{lee2015m}.
Dropout \cite{srivastava2014dropout} can be interpreted as an ensemble approach which takes exponential number of networks with high correlation.
In addition to dropout, Veit~\etal~\cite{veit2016residual} state that \emph{residual networks} behave like ensembles of relatively shallow networks.
Recently the ensemble technique has been applied in deep metric learning as well.
Yuan \etal~\cite{yuan2016hard} propose to ensemble a set of models with different complexities in cascaded manner.
They train deeply supervised cascaded networks using easier examples through earlier layers of the networks while harder examples are further exploited in later layers.
Opitz \etal~\cite{opitz2017bier} use online gradient boosting to train each learner in ensemble.
They try to reduce correlation among learners using re-weighting of training samples.
% depending on the loss function.
%contrast to \cite{yuan2016hard} which uses sampling for hard training examples.
Opitz \etal~\cite{opitz2016efficient} propose an efficient averaging strategy with a novel \emph{DivLoss} which encourages diversity of individual learners.

\vspace{-5mm}
\subsubsection{Attention mechanism}
\label{sec:related-attention}
Attention mechanism has been used in various computer vision problems.
Earlier researches utilize RNN architectures for attention modeling~\cite{sermanet2014attention,mnih2014recurrent,ba2014multiple}.
These RNN based attention models solve classification tasks using object parts detection by sequentially selecting attention regions from images and then learning feature representations for each part.
%parts expressions from each region.
%%To optimize these RNN based networks, however, policy gradient estimation is required contrast to our proposed method which is in a fully gradient-based way.
% done so that training cannot be performed in end-to-end manner.
Besides RNN approaches, Liu \etal~\cite{liu2016fully} propose \emph{fully convolutional attention networks}, which adopts hard attention from a region generator.
And Zhao \etal~\cite{zhao2017diversified} propose \emph{diversified visual attention networks}, which uses different scaling or cropping of input images for different attention masks.
%proposed by , the other is \emph{diversified visual attention networks} by Zhao \etal~\cite{zhao2017diversified}.
%Besides RNN approaches, two works are closely related to our ABE-$M$; the one is \emph{fully convolutional attention networks} proposed by Liu \etal~\cite{liu2016fully}, the other is \emph{diversified visual attention networks} by Zhao \etal~\cite{zhao2017diversified}.
%To attend to multiple regions of an object, \cite{liu2016fully} adopts hard attention from a region generator and \cite{zhao2017diversified} uses different scaling or cropping of input images for different attention masks.
%, which attends to multiple parts of an object.
%However, our proposed attention model branches from multiple layers while their attention model branches from a fixed layer.
%Also, \cite{liu2016fully} requires each attention branch to have its own layers (and associated weights and gradient values for back-propagation) but our model shares single attention layer for each branch which is more efficient in both training and testing.
%Zhao \etal~\cite{zhao2017diversified} proposed attention-based networks which is relevant to our work.
%, however, their attention region generation is non-differentiable.
%However, there is a drawback of these approaches that they are non-differentiable.
However, our ABE-$M$ is able to learn diverse attention masks without relying on a region generator.
% for every attention masks.
% while \cite{zhao2017diversified} uses different scaling or cropping of input images for different attention masks.
In addition, ABE-$M$ uses soft attention, therefore, the parameter update is straightforward by backpropagation in a fully gradient-based way while previous approaches in \cite{sermanet2014attention,mnih2014recurrent,ba2014multiple,liu2016fully,zhao2017diversified} use hard attention which requires policy gradient estimation.

Jaderberg \etal~\cite{jaderberg2015spatial} propose spatial transformer networks which models attention mechanism using parameterized image transformations.
Unlike aforementioned approaches, their model is differentiable and thus can be trained in a fully gradient-based way.
%Unlike other approaches which are non-differentiable, their model is clearly differentiable and even can learn multiple learners by simulating multiple \emph{transformers}.
However, their attention is limited to a set of predefined and parameterized transformations which could not yield arbitrary attention masks.
%there is a limitation that it requires pre-defined transformations and cannot generate arbitrary attention regions.

%% file: proposed-method.tex
\vspace{-4mm}
\section{Attention-based ensemble}
\label{sec:proposed}
\vspace{-2mm}
\subsection{Deep metric learning}
\vspace{-2mm}

Let $f:\mathcal{X} \rightarrow \mathcal{Y}$ be an isometric embedding function
 between metric spaces $\mathcal{X} $ and $\mathcal{Y}$
 where $\mathcal{X} $ is a $N_\mathcal{X}  $ dimensional metric space with an unknown metric function $d_\mathcal{X} $ and
 $\mathcal{Y}$ is a $N_\mathcal{Y}$ dimensional metric space with a known metric function $d_\mathcal{Y} $.
For example, $\mathcal{Y}$ could be a Euclidean space with Euclidean distance or the unit sphere in a Euclidean space with angular distance.

Our goal is to approximate $f$ with a deep neural network
 from a dataset $\mathcal{D}=\{(x^{(1)},x^{(2)},d_\mathcal{X}  (x^{(1)},x^{(2)} ))|x^{(1)},x^{(2)}\in \mathcal{X} \}$
 which are samples from $\mathcal{X} $. In case we cannot get the samples of metric $d_\mathcal{X} $,
 we consider the label information from the dataset with labels as the relative constraint of the metric $d_\mathcal{X} $.
For example, from a dataset $\mathcal{D}_\mathcal{C}=\{(x,c)|x\in \mathcal{X} ,c\in \mathcal{C}\}$ where $\mathcal{C}$ is the set of labels,
     for $(x_i,c_i ),(x_j,c_j )\in \mathcal{D}_\mathcal{C} $
 the contrastive metric constraint could be defined as the following:
%\vspace{-2mm}
\begin{equation}
 \begin{cases}
 d_\mathcal{X} (x_i,x_j ) =0 , &\text{if $c_i=c_j $; } \\
 d_\mathcal{X} (x_i,x_j )>m_c, &\text{if $c_i\neq c_j$,}
 \end{cases}
\vspace{-1mm}
\label{eq:contrastive_metric}
\end{equation}
where $m_c$ is an arbitrary margin.
The triplet metric constraint
 for $(x_i,c_i ),$ $(x_j,c_j ),$ $(x_k,c_k )$ $\in \mathcal{D}_\mathcal{C} $
 could be defined as the following:
\vspace{-2mm}
\begin{equation}
   d_\mathcal{X} (x_i,x_j) +m_t <d_\mathcal{X}  (x_i,x_k), \text{~~$c_i=c_j $ and $c_i \neq c_k $},
\label{eq:triplet_metric}
\vspace{-3mm}
\end{equation}
 where $m_t$ is a margin.
Note that these metric constraints are some choices of how to model $d_\mathcal{X} $, not those of how to model $f$.

An embedding function $f$ is isometric or distance preserving embedding if for every $x_i,x_j \in \mathcal{X}$ one has $d_\mathcal{X}  (x_i,x_j )=d_\mathcal{Y} (f(x_i),f(x_j) )$.
In order to have an isometric embedding function $f$, 
we optimize $f$ so that the points embedded into $\mathcal{Y}$ produce exactly the same metric or obey the same metric constraint of $d_\mathcal{X}$.

\vspace{-4mm}
\subsection{Ensemble for deep metric learning}
\label{sec:ensemble_dml}
\vspace{-1mm}

A classical ensemble for deep metric learning could be the method to average the metric of multiple embedding functions.
We define the ensemble metric function $d_{\mathrm{ensemble}} $ for deep metric learning as the following:
\vspace{-3mm}
\begin{equation}
d_{\mathrm{ensemble},{(f_1,\dots,f_{\Ensemblesub})}}  (x_i,x_j )=\frac{1}{\Ensemblesub} \sum_{\ensemblesub=1}^{\Ensemblesub} d_\mathcal{Y} (f_\ensemblesub (x_i ),f_\ensemblesub (x_j)),
\label{eq:d_ensemble}
\vspace{-2mm}
\end{equation}
where $f_\ensemblesub$ is an independently trained embedding function and we call it a learner.

In addition to the classical ensemble, we can consider the ensemble of two-step embedding function.
Consider a function $s:\mathcal{X}  \rightarrow \mathcal{Z}$ which is an isometric embedding function
 between metric spaces $\mathcal{X} $ and $\mathcal{Z}$ where $\mathcal{X} $ is a $N_\mathcal{X}$
 dimensional metric space with an unknown metric function $d_\mathcal{X}  $ and $\mathcal{Z}$
 is a $N_\mathcal{Z} $ dimensional metric space with an unknown metric function $d_\mathcal{Z}$.
And we consider the isometric embedding $g:\mathcal{Z} \rightarrow \mathcal{Y}$ where
 $\mathcal{Y}$ is a $N_\mathcal{Y}$ dimensional metric space with a known metric function $d_\mathcal{Y}$.
If we combine them into one function $\ensemblefunc(x)=g(s(x)),x \in \mathcal{X} $,
 the combined function is also an isometric embedding $\ensemblefunc:\mathcal{X} \rightarrow \mathcal{Y}$
 between metric spaces $\mathcal{X} $ and $\mathcal{Y}$.

Like the parameter sharing ensemble \cite{lee2015m}, with the independently trained multiple
 $g_\ensemblesub$ and a single $s$, we can get multiple embedding functions $\ensemblefunc_\ensemblesub:\mathcal{X} \rightarrow \mathcal{Y}$ as the following:
\vspace{-2mm}
\begin{equation}
\ensemblefunc_\ensemblesub (x)=g_\ensemblesub (s(x)) .
\label{eq:multihead}
\vspace{-2mm}
\end{equation}

We are interested in another case where there are multiple embedding functions $\ensemblefunc_\ensemblesub:\mathcal{X}  \rightarrow \mathcal{Y}$ with multiple $s_\ensemblesub$ and a single $g$ as the following:
\vspace{-2mm}
\begin{equation}
\ensemblefunc_\ensemblesub (x)=g(s_\ensemblesub (x)) .
\vspace{-2mm}
\label{eq:m_k}
\end{equation}

Note that a point in $\mathcal{X} $ can be embedded into multiple points in $\mathcal{Y}$ by multiple learners.
In Eq. (\ref{eq:m_k}), $s_\ensemblesub$ does not have to preserve the label information while it only has to preserve the metric.
In other words, a point with a label could be mapped to multiple locations in $\mathcal{Z}$ by multiple $s_\ensemblesub$ and
 finally would be mapped to multiple locations in $\mathcal{Y}$.
If this were the ensemble of classification models where $g$ approximates the distribution of the labels,
 all $s_\ensemblesub$ should be label preserving functions because the outputs of $s_\ensemblesub$
 become the inputs of one classification model $g$.

For the embedding function of Eq. (\ref{eq:m_k}), we want to make $s_\ensemblesub$ attends to the diverse aspects of data $x$ in $\mathcal{X}$
 while maintaining a single embedding function $g$ which disentangles the complex manifold $\mathcal{Z}$ into Euclidean space.
By exploiting the fact that a point $x$ in $\mathcal{X}$ can be mapped to multiple locations in $\mathcal{Y}$,
 we can encourage each $s_\ensemblesub$ to map $x$ into distinctive points $z_\ensemblesub$ in $\mathcal{Z}$.
Given an isometric embedding $g:\mathcal{Z} \rightarrow \mathcal{Y}$ ,
 if we enforce $y_\ensemblesub$ in $\mathcal{Y}$ mapped from $x$ to be far from each other,
 $z_\ensemblesub$ in $\mathcal{Z}$ mapped from $x$ will be far from each other as well.
Note that we cannot apply this divergence constraint to $z_\ensemblesub$ because metric $d_z$ in $\mathcal{Z}$ is unknown.
 We train each $\ensemblefunc_\ensemblesub $ to be isometric function between $\mathcal{X} $ and $\mathcal{Y}$
 while applying the divergence constraint among $y_\ensemblesub$ in $\mathcal{Y}$.
If we apply the divergence constraint to classical ensemble models or multihead ensemble models,
 they do not necessarily induce the diversity because each $f_\ensemblesub $ or $g_\ensemblesub$ could
 arbitrarily compose different metric spaces in $\mathcal{Y}$ (Refer to experimental results in Sec. \ref{sec:effectsofdivloss}).
With the attention-based ensemble, union of metric spaces by multiple $s_\ensemblesub $ is mapped by a single embedding function $g$.

\vspace{-4mm}
\subsection{Attention-based ensemble model}
\label{sec:model}
\vspace{-2mm}

\begin{figure}[t]
\centering
\includegraphics[width=0.9\linewidth]{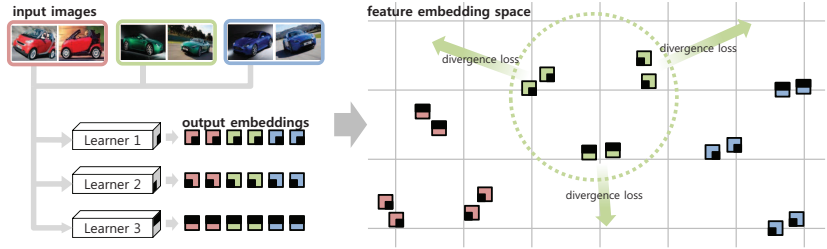}
\vspace{-3mm}
\caption{Illustration of feature embedding space and divergence loss. Different car brands are represented as different colors: red, green and blue. Feature embeddings of each learner are depicted as a square with different mask patterns. Divergence loss pulls apart the feature embeddings of different learners using same input}
\vspace{-5mm}
\label{fig:loss}
\end{figure}

As one implementation of Eq.(\ref{eq:m_k}), we propose the attention-based ensemble model which is mainly composed of two parts:
 feature extraction module $F(x)$ and attention module $A(x)$. For the feature extraction, we assume a general multi-layer perceptron model as the following:
\vspace{-2mm}
\begin{equation}
	F(x)=h_l(h_{l-1}( \cdots (h_2(h_1(x))))
\label{eq:feature}
\vspace{-2mm}
\end{equation}

We break it into two parts with a branching point at $i$, $S(\cdot)$ includes $h_l$, $h_{l-1}, \dots, h_{i+1}$,
 and $G(\cdot)$ includes $h_i$, $h_{i-1}$, $\dots$, $h_1$.
We call $S(\cdot)$ a spatial feature extractor and $G(\cdot)$ a global feature embedding function
 with respect to the output of each function. For attention module,
 we also assume a general multi-layer perceptron model which outputs a three dimensional blob with channel, width, and height as an attention mask.
Each element in the attention masks is assumed to have a value from 0 to 1.
Given aforementioned two modules, the combined embedding function $\Ensemblefunc_\ensemblesub(x)$ for the learner $\ensemblesub$ is defined as the following:
\vspace{-2mm}
\begin{equation}
	\Ensemblefunc_\ensemblesub(x) = G(S(x) \circ A_\ensemblesub(S(x)) ),
\label{eq:embedding}
\vspace{-2mm}
\end{equation}
where $\circ$ denotes element-wise product (Fig. \ref{fig:attentionbased_ensemble}).

Note that, same feature extraction module is shared across different learners while individual learners have their own attention module $A_\ensemblesub(\cdot)$.
The attention function $A_\ensemblesub(S(x))$ outputs an attention mask with same size as output of $S(x)$.
This attention mask is applied to the output feature of $S(x)$ with an element-wise product.
Attended feature output of $S(x) \circ A_\ensemblesub(S(x))$ is then fed into global feature embedding function
 $G(\cdot)$ to generate an embedding feature vector.
If all the elements in the attention mask are 1, the model $\Ensemblefunc_\ensemblesub(x)$ is reduced to a conventional multi-layer perceptron model.

\vspace{-4mm}
\subsection{Loss}
\label{sec:loss}
\vspace{-2mm}

The loss for training aforementioned attention model is defined as:

\vspace{-2mm}
\begin{equation}
	L(\{(x_i,c_i)\}) = \sum_\ensemblesub L_{\mathrm{metric}, (\ensemblesub)}(\{(x_i,c_i)\}) + \lambda_{\mathrm{div}} L_{\mathrm{div}}(\{x_i\}),
\vspace{-1mm}
\label{eq:loss}
\end{equation}
where $\{(x_i,c_i)\}$ is a set of all training samples and labels, $L_{\mathrm{metric}, (\ensemblesub)}(\cdot)$ is the loss for the isometric embedding for the $\ensemblesub$-th learner, $L_{\mathrm{div}}(\cdot)$ is regularizing term for diversifying the feature embedding of each learner $\Ensemblefunc_\ensemblesub(x)$ and $\lambda_{\mathrm{div}}$ is the weighting parameter to control the strength of the regularizer. More specifically, divergence loss $L_{\mathrm{div}}$ is defined as the following:

\vspace{-2mm}
\begin{equation}
	L_{\mathrm{div}}(\{x_i\}) = \sum_i \sum_{p,q} \max(0, m_{\mathrm{div}} - d_{\mathcal{Y}}(\Ensemblefunc_p(x_i), \Ensemblefunc_q(x_i))^2) \vspace{-1mm}
\label{eq:div_loss},
\end{equation}
where $\{x_i\}$ is set of all training samples, $d_{\mathcal{Y}}$ is the metric in $\mathcal{Y}$ and $m_{\mathrm{div}}$ is a margin.
A pair $(\Ensemblefunc_p(x_i), \Ensemblefunc_q(x_i))$ represents feature embeddings of a single image embedded by two different learners.
We call it self pair from now on
while positive and negative pairs refer to pairs of feature embeddings with same labels and different labels, respectively.

The divergence loss encourages each learner to attend to the different part of the input image by increasing the distance between the points embedded by the input image (Fig. \ref{fig:loss}).
Since the learners share the same functional module to extract features, the only differentiating part is the attention module.
Note that our proposed loss is not directly applied to the attention masks.
In other words, the attention masks among the learners may overlap.
And also it is possible to have the attention masks some of which focus on small region while other focus on larger region including small one.

%% file: implementation.tex
\vspace{-4mm}
\section{Implementation}
\label{sec:implementation}
\vspace{-2mm}

\begin{figure}[t]
\centering
\includegraphics[width=0.9\linewidth]{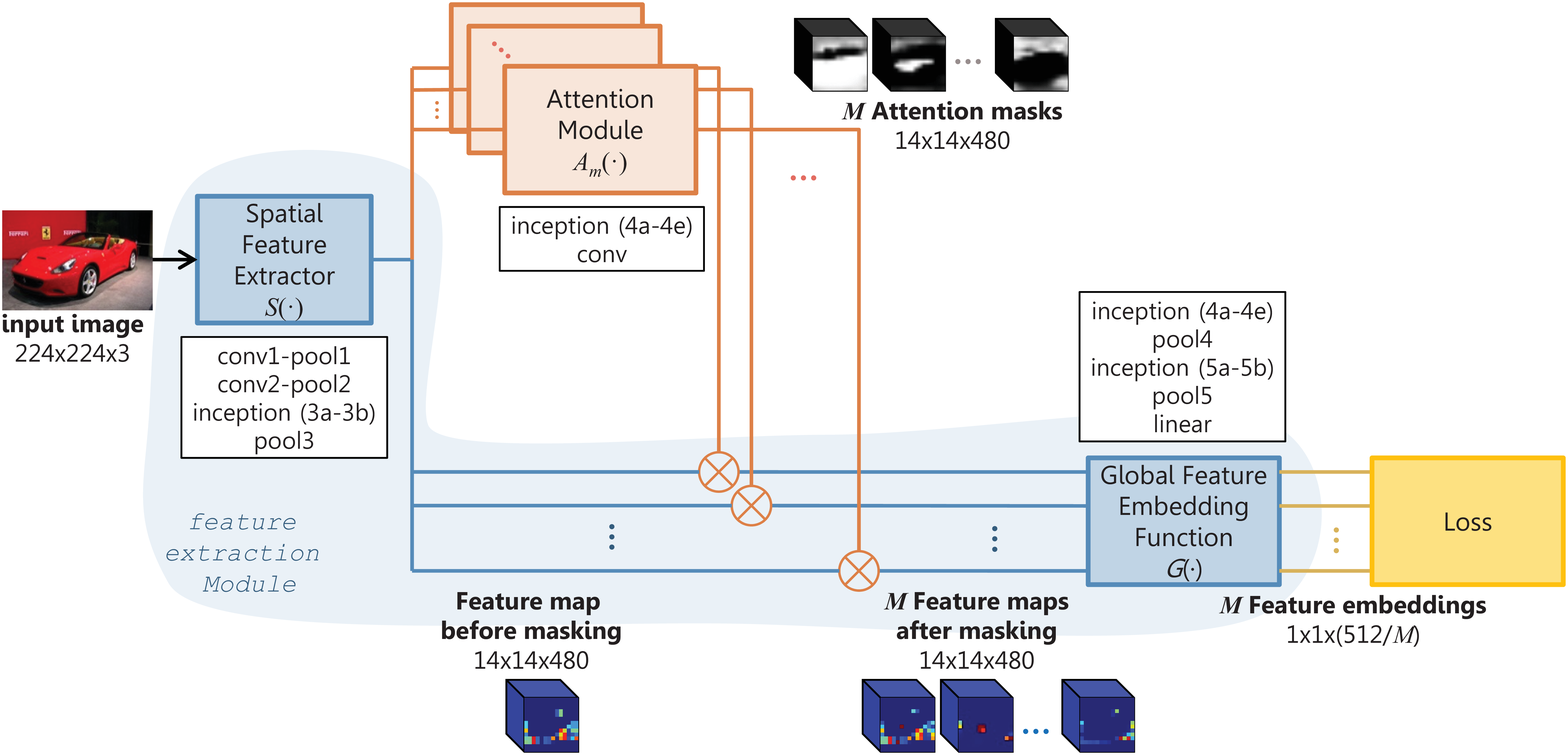}
\vspace{-4mm}
\caption{The implementation of attention-based ensemble (ABE-$M$) using GoogLeNet}
\label{fig:architecture}
\vspace{-3mm}
\end{figure}

We perform all our experiments using GoogLeNet \cite{Szegedy_2015_CVPR} as the base architecture. 
As shown in Fig.~\ref{fig:architecture}, 
we use the output of max pooling layer following the \texttt{inception(3b)} block as our spatial feature extractor $S(\cdot)$ and remaining network as our global feature embedding function $G(\cdot)$.
In our implementation, we simplify attention module $A_\ensemblesub(\cdot)$ as $A_\ensemblesub'(C(\cdot))$ where $C(\cdot)$ consists of \texttt{inception(4a)} to \texttt{inception(4e)} from GoogLeNet, which is shared among all $M$ learners and $A_\ensemblesub'(\cdot)$ consists of a convolution layer of 480 kernels of size 1$\times$1 to match the output of $S(\cdot)$ for the element-wise product.
This is for efficiency
in terms of memory and computation time. Since $C(\cdot)$ is shared across
different learners, forward and backward propagation time,
memory usage, and number of parameters are decreased compared to having separate $A_\ensemblesub(\cdot)$ for each learner
(without any shared part). Our preliminary experiments showed no performance drop with this choice of implementation.

We study the effects of different branching points and depth of attention module in Sec. \ref{sec:ablation}.
We use contrastive loss \cite{hadsell2006dimensionality,bell2015learning,chopra2005learning} as our distance metric loss function which is defined as the following:
\vspace{-2mm}
\begin{equation}
\begin{aligned}
L_{\mathrm{metric}, (\ensemblesub)}(\{(x_i,c_i)\}) &  = \frac{1}{N}\sum_{i,j} (1-y_{i,j})[m_{c}-D_{\ensemblesub,i,j}^2]_+ + y_{i,j} D_{\ensemblesub,i,j}^2, \\
D_{\ensemblesub,i,j} & =d_{\mathcal{Y}}(\Ensemblefunc_\ensemblesub(x_i), \Ensemblefunc_\ensemblesub(x_j)),
\end{aligned}
\label{eq:metricloss}
\vspace{-2mm}
\end{equation}
where $\{(x_i,c_i)\}$ is set of all training samples and corresponding labels, $N$ is the number of training sets, $y_{i,j}$ is a binary indicator of whether or not the label $c_i$ is equal to $c_j$, $d_{\mathcal{Y}}$ is the euclidean distance, $[\cdot]_+$ denotes the hinge function $\max(0,\cdot)$ and $m_{c}$ is the margin for contrastive loss. Both of margins $m_{c}$ and $m_\mathrm{div}$ (in Eq. \ref{eq:loss}) is set to 1.

We implement the proposed ABE-$M$ method using caffe \cite{jia2014caffe} framework.
During training, the network is initialized from a pre-trained network on ImageNet ILSVRC dataset \cite{russakovsky2015imagenet}.
The final layer of the network and the convolution layer of attention module are randomly initialized as proposed by Glorot~\etal~\cite{glorot2010understanding}.
For optimizer, we use stochastic gradient descent with momentum optimizer with momentum as 0.9, and we select the base learning rate by tuning on validation set of the dataset.

We follow earlier works \cite{oh2016deep,wu2017sampling} for preprocessing and unless stated otherwise, we use the input image size of 224$\times$224.
All training and testing images are scaled such that their longer side is 256, keeping the aspect ratio fixed, and padding the shorter side to get 256$\times$256 images. During training, we randomly crop images to 224$\times$224 and then randomly flip horizontally.
During testing, we use the center crop. We subtract the channel-wise mean of ImageNet dataset from the images.
For training and testing images of cropped datasets, we follow the approach in \cite{wu2017sampling}.
For CARS-196 \cite{KrauseStarkDengFei-Fei_3DRR2013} cropped dataset, 256$\times$256 scaled cropped images are used; while for CUB-200-2011 \cite{WahCUB_200_2011} cropped dataset, 256$\times$256 scaled cropped images with fixed aspect ratio and shorter side padded are used.

We run our experiments on nVidia Tesla M40 GPU (24GBs GPU memory), which limits our batch size to 64 for ABE-$8$ model.
Unless stated otherwise, we use the batch size of 64 for our experiments.
We sample our mini-batches by first randomly sampling 32 images and then positive pairs for first 16 images and negative pairs for next 16 images, thus making the mini-batch of size 64.
Unless mentioned otherwise, we report the results of our method using embedding size of 512. This makes the embedding size for individual learners to be 512/$M$.

%% file: evaluation.tex
\vspace{-3mm}
\section{Evaluation}
\vspace{-3mm}
We use all commonly used image retrieval task datasets for our experiments and Recall@$K$ metric for our evaluation.
During testing, we compute the feature embeddings for all the test images from our network.
For every test image, we then retrieve top $K$ similar images from the test set excluding test image itself.
Recall score for that test image is 1 if at least one image out of $K$ retrieved images has the same label as the test image.
We compute the average over whole test set to get Recall@$K$.
We evaluate the model after every 1000 iteration and report the results for the iteration with highest Recall@1.

We show the effectiveness of the proposed ABE-$M$ method on all the datasets commonly used in image retrieval tasks. We follow same train-test split as \cite{oh2016deep} for fair comparison with other works.
\begin{itemize}
\item \textbf{CARS-196} \cite{KrauseStarkDengFei-Fei_3DRR2013} dataset contains images of 196 different classes of cars and is primarily used for our experiments. The dataset is split into 8,144 training images and 8,041 testing images (98 classes in both).
\item \textbf{CUB-200-2011} \cite{WahCUB_200_2011} dataset consists of 11,788 images of 200 different bird species. We use the first 100 classes for training (5,864 images) and the remaining 100 classes for testing (5,924 images).
\item \textbf{Stanford online products (SOP)} \cite{oh2016deep} dataset has 22,634 classes with 120,053 product images. 11,318 classes are used for training (59,551 images) while other 11,316 classes are for testing (60,502 images).
\item \textbf{In-shop clothes retrieval} \cite{liu2016deepfashion} dataset contains 11,735 classes of clothing items with 54,642 images. Following similar protocol as \cite{oh2016deep}, we use 3,997 classes for training (25,882 images) and other 3,985 classes for testing (28,760 images). The test set is partitioned into the query set of 3,985 classes (14,218 images) and the retrieval database set of 3,985 classes (12,612 images).
\end{itemize}
Since CARS-196 and CUB-200-2011 datasets consist of bounding boxes too, we report the results using original images and cropped images both for fair comparison. \iffalse Following the approach in \cite{wu2017sampling}, for CARS196 cropped dataset, we scale the cropped images to 256x256; and for CUB200 cropped, we scale and pad the images such that their longer side is 256 pixels, keeping the aspect ratio fixed.\fi

%% file: experiments.tex
\vspace{-2mm}
\section{Experiments} \label{experiments}
\vspace{-2mm}
\subsection{Comparison of ABE-$M$ with $M$-heads}
\vspace{-2mm}

\begin{figure}[t]
\begin{center}
    \mbox{%
    \subfigure[]{ \label{fig:r_vs_param} \includegraphics[width=0.3\linewidth]{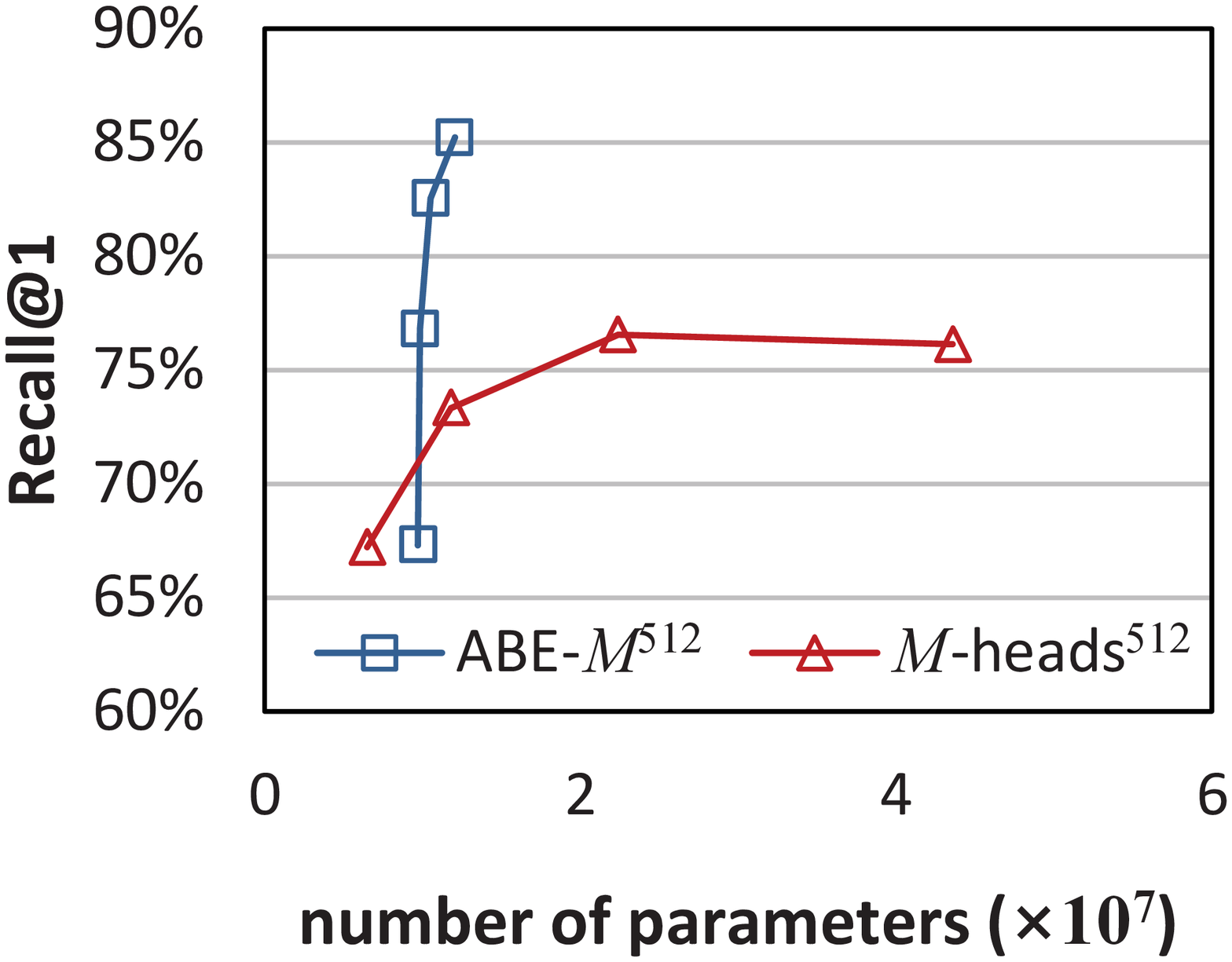}}
%    }
\hspace{5mm}
 %   \mbox{%
    \subfigure[]{ \label{fig:r_vs_flops} \includegraphics[width=0.3\linewidth]{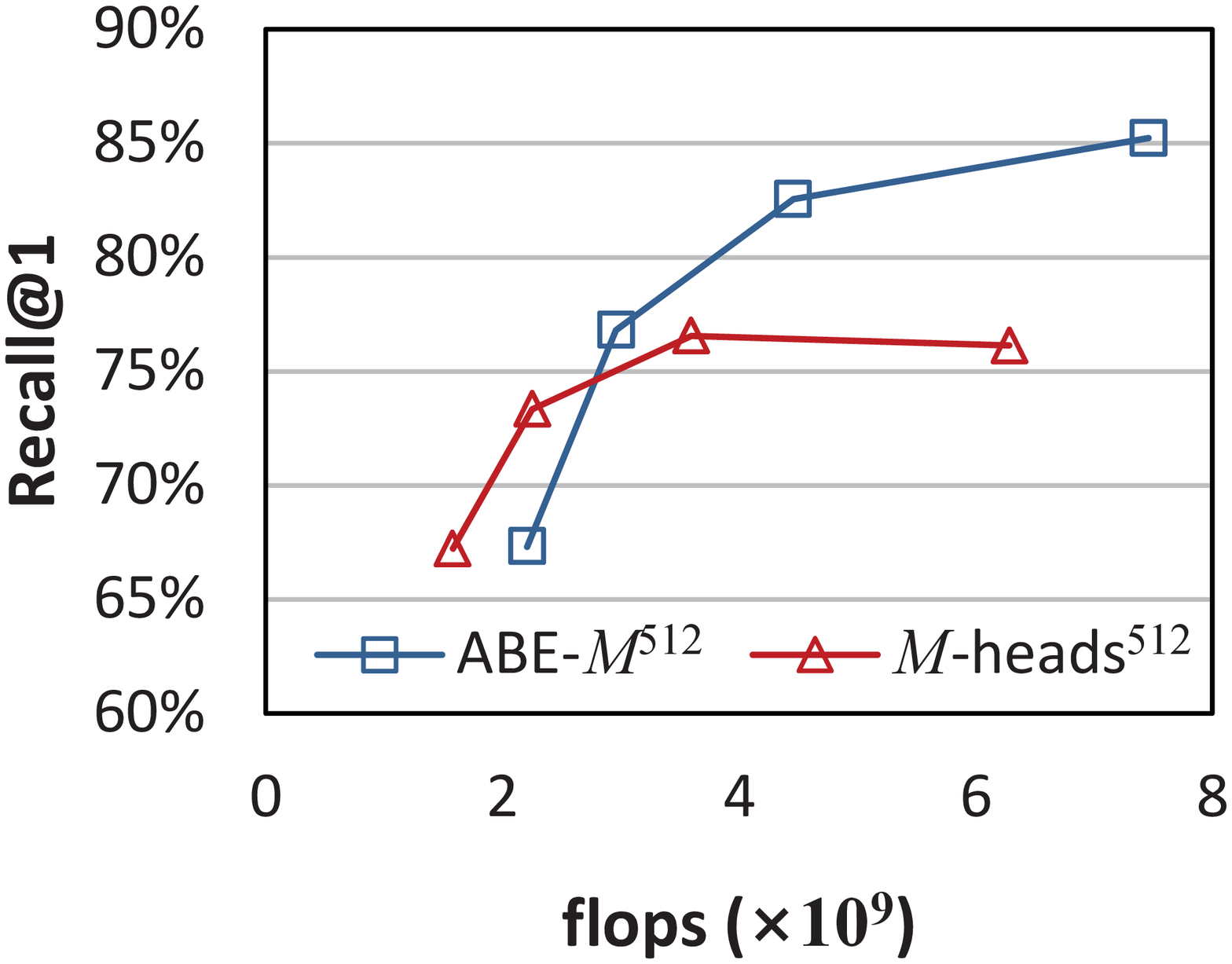}}
    }
\end{center}
\vspace{-8mm}
\caption{Recall@1 comparison with baseline on CARS-196 as a function of (a) number of parameters and (b) flops. Both of ABE-$M$ and $M$-heads has embedding size of 512}
\label{fig:r_vs_x}
\vspace{-3mm}
\end{figure}

To show the effectiveness of our ABE-$M$ method, we first compare the performance of ABE-$M$ and $M$-heads ensemble (Fig.~\ref{fig:conventional_ensemble}) with varying ensemble embedding sizes (denoted with superscript) on CARS-196 dataset.
As show in Table~\ref{table:baselinecomptable} and Fig. \ref{fig:r_vs_x},
our method outperforms $M$-heads ensemble by a significant margin.
The number of model parameters for ABE-$M$ is much less compared to $M$-heads ensemble as the global feature extractor $G(\cdot)$ is shared among learners.
But, ABE-$M$ requires higher flops because of extra computation of attention modules. This difference becomes increasingly insignificant with increasing values of $M$.

ABE-1 contains only one attention module and hence is not an ensemble and does not use divergence loss.
ABE-1 performs similar to 1-head.
We also report the performance of individual learners of the ensemble.
From Table \ref{table:baselinecomptable}, we can see that the performance of ABE-$M^{512}$ ensemble is increasing with increasing $M$.
The performance of individual learners is also increasing with increasing $M$ despite the decrease in embedding size of individual learners (512/$M$).
The same increase is not seen for the case of $M$-heads.
Further, we can refer to ABE-1$^{64}$, ABE-2$^{128}$, ABE-4$^{256}$ and ABE-8$^{512}$, where all individual learners have embedding size 64.
We can see a clear increase in recall of individual learners with increasing values of $M$.

%From Table \ref{table:baselinecomptable}, we can see that the performance of individual learners is degrading with increasing $M$ in the case of $M$-heads.
%We should note that embedding size of individual learners (512/$M$) is also decreasing in this case.
%For ABE-$M$$^{512}$, we see an increase in the recall of individual models despite the decrease in embedding size.
%Further, we can refer to ABE-1$^{64}$, ABE-2$^{128}$, ABE-4$^{256}$ and ABE-8$^{512}$, where all individual learners have embedding size 64.
%We can see a clear increase in recall of individual learners with increasing values of $M$.
%This shows that ABE-$M$ not only increases the performance of the ensemble but also the individual learners.

\setlength{\tabcolsep}{4pt}
\begin{table}
\begin{center}
%\vspace{-2mm}
\caption{Recall@$K$(\%) comparison with baseline on CARS-196. Superscript denotes ensemble embedding size}
\vspace{-3mm}
\label{table:baselinecomptable}
%\fontsize{9}{10}\selectfont
\scriptsize
\begin{tabular}{Sl   c c c c      c     c c c c    c c }
\hline
 & \multicolumn{4}{c}{{Ensemble}} && \multicolumn{4}{c}{{Individual Learners}} & \tiny{params} &  \tiny{flops} \\
\cline{2-5}
\cline{7-10}
\hspace{-1mm}$K$ & 1 & 2 & 4 & 8 & & 1 & 2 & 4 & 8  & \tiny{($\times 10^7$)}  &  \tiny{($\times 10^9$)}   \\
\hline
\hspace{-1mm}1-head$^{512}$ & 67.2 & 77.4 & 85.3 & 90.7 && - & - &- &-- & 0.65 & 1.58\\
\hspace{-1mm}2-heads$^{512}$ & 73.3 & 82.5 & 88.6 & 93.0 && 70.2\tiny{$\pm$.03} & 79.8\tiny{$\pm$.52} &86.7\tiny{$\pm$.01} &91.9\tiny{$\pm$.37}&1.18&2.25\\
\hspace{-1mm}4-heads$^{512}$ & 76.6 & 84.2 & 89.3 & 93.2 && 70.4\tiny{$\pm$.80} & 79.9\tiny{$\pm$.38} &86.5\tiny{$\pm$.43} &91.4\tiny{$\pm$.42}&2.24&3.60\\
\hspace{-1mm}8-heads$^{512}$ & 76.1 & 84.3 & 90.3 & 93.9 && 68.3\tiny{$\pm$.39} & 78.5\tiny{$\pm$.39} &86.0\tiny{$\pm$.37} &91.3\tiny{$\pm$.31}&4.36&6.28\\
\hspace{-1mm}ABE-1$^{512}$ & 67.3 & 77.3 & 85.3 & 90.9 && - & - &- &- & 0.97 & 2.21\\
\hspace{-1mm}ABE-2$^{512}$ & 76.8 & 84.9 & 90.2 & 94.0 && 70.9\tiny{$\pm$.58} & 80.3\tiny{$\pm$.04} &87.1\tiny{$\pm$.07} &92.2\tiny{$\pm$.20} & 0.98 & 2.96\\
\hspace{-1mm}ABE-4$^{512}$ & \underline{82.5} & \underline{89.1} & \underline{93.0} & \underline{95.5} && 74.4\tiny{$\pm$.51} & 83.1\tiny{$\pm$.47} &89.1\tiny{$\pm$.34} &93.2\tiny{$\pm$.36} & 1.05 & 4.46\\
\hspace{-1mm}ABE-8$^{512}$ & \textbf{85.2} & \textbf{90.5} & \textbf{93.9} & \textbf{96.1} && 75.0\tiny{$\pm$.39} & 83.4\tiny{$\pm$.24} &89.2\tiny{$\pm$.31} &93.2\tiny{$\pm$.24} & 1.20 & 7.46\\
\hspace{-1mm}ABE-1$^{64}$ & 65.9 & 76.5 & 83.7 & 89.3 && - & - &- &- & 0.92 & 2.21\\
\hspace{-1mm}ABE-2$^{128}$ & 75.5 & 84.0 & 89.4 & 93.6 && 68.6\tiny{$\pm$.38} & 78.8\tiny{$\pm$.38} &85.7\tiny{$\pm$.43} &91.3\tiny{$\pm$.16} & 0.96 & 2.96\\
\hspace{-1mm}ABE-4$^{256}$ & 81.8 & 88.5 & 92.4 & 95.1 && 72.3\tiny{$\pm$.68} & 81.4\tiny{$\pm$.45} &87.9\tiny{$\pm$.23} &92.3\tiny{$\pm$.13} & 1.04 & 4.46\\
\hline
\end{tabular}
\end{center}
\end{table}
\label{table:baseline}
\vspace{-6mm}
\setlength{\tabcolsep}{1.4pt}

\vspace{-3mm}
\subsection{Effects of divergence loss}\label{sec:effectsofdivloss}
\vspace{-1mm}

\begin{figure}[t]
\begin{center}
    \mbox{%
    \subfigure[]{ \label{fig:histogram_a} \includegraphics[width=0.22\linewidth]{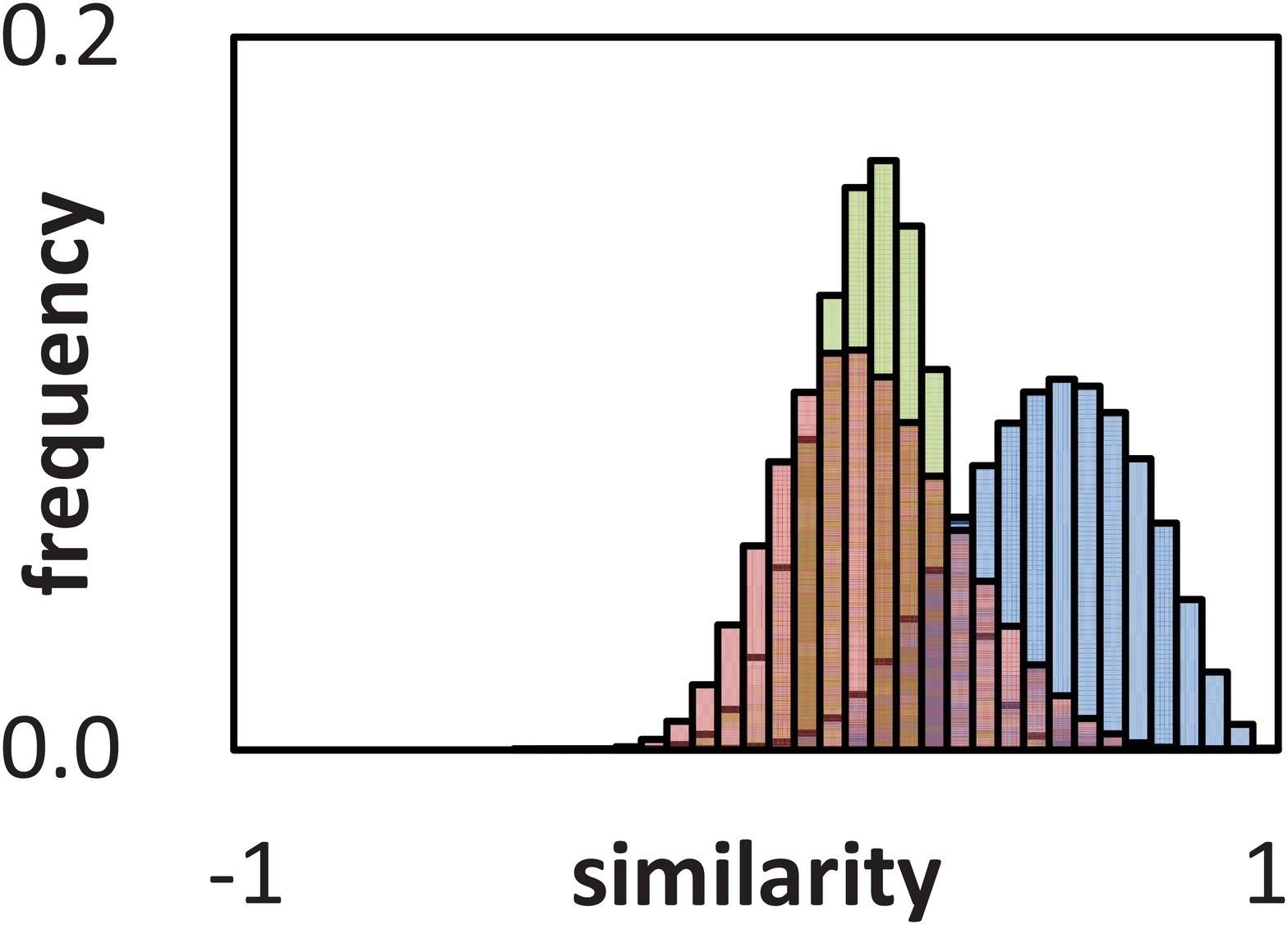}}
%    }
 %   \mbox{%
    \subfigure[]{ \label{fig:histogram_b} \includegraphics[width=0.22\linewidth]{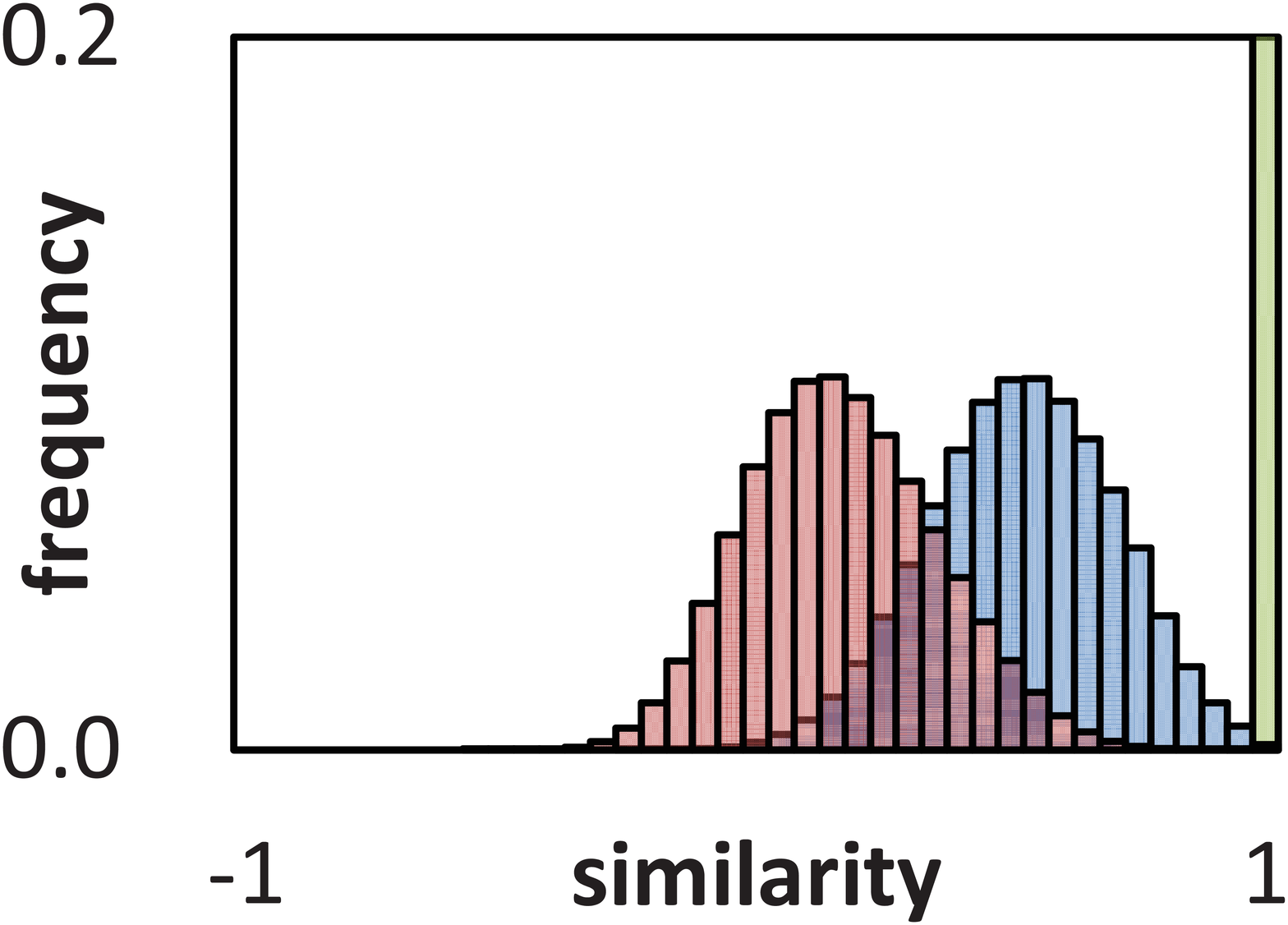}}
%    }
%    \mbox{%
    \subfigure[]{ \label{fig:histogram_c} \includegraphics[width=0.22\linewidth]{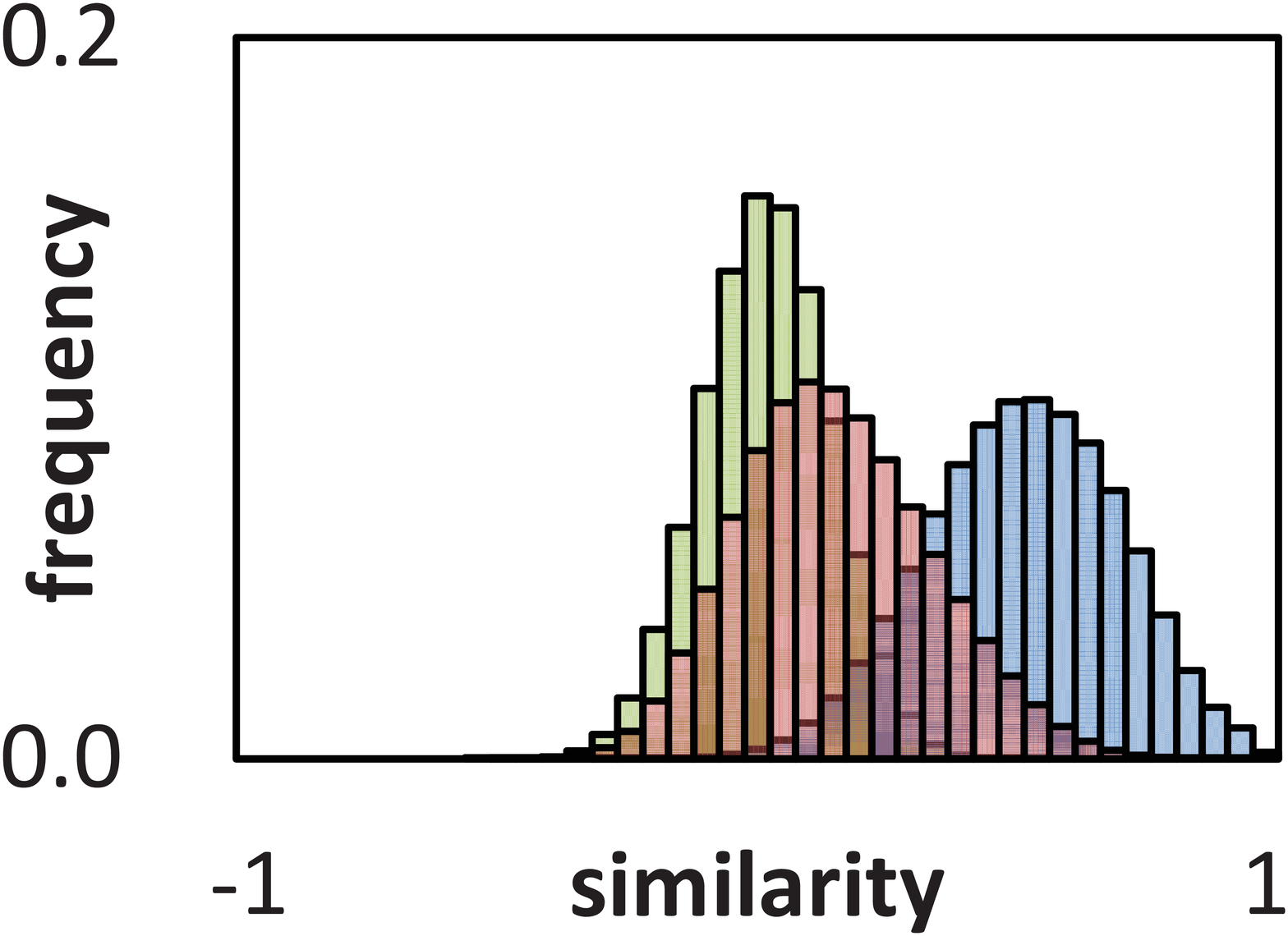}}
%    }
 %   \mbox{%
    \subfigure[]{ \label{fig:histogram_d} \includegraphics[width=0.22\linewidth]{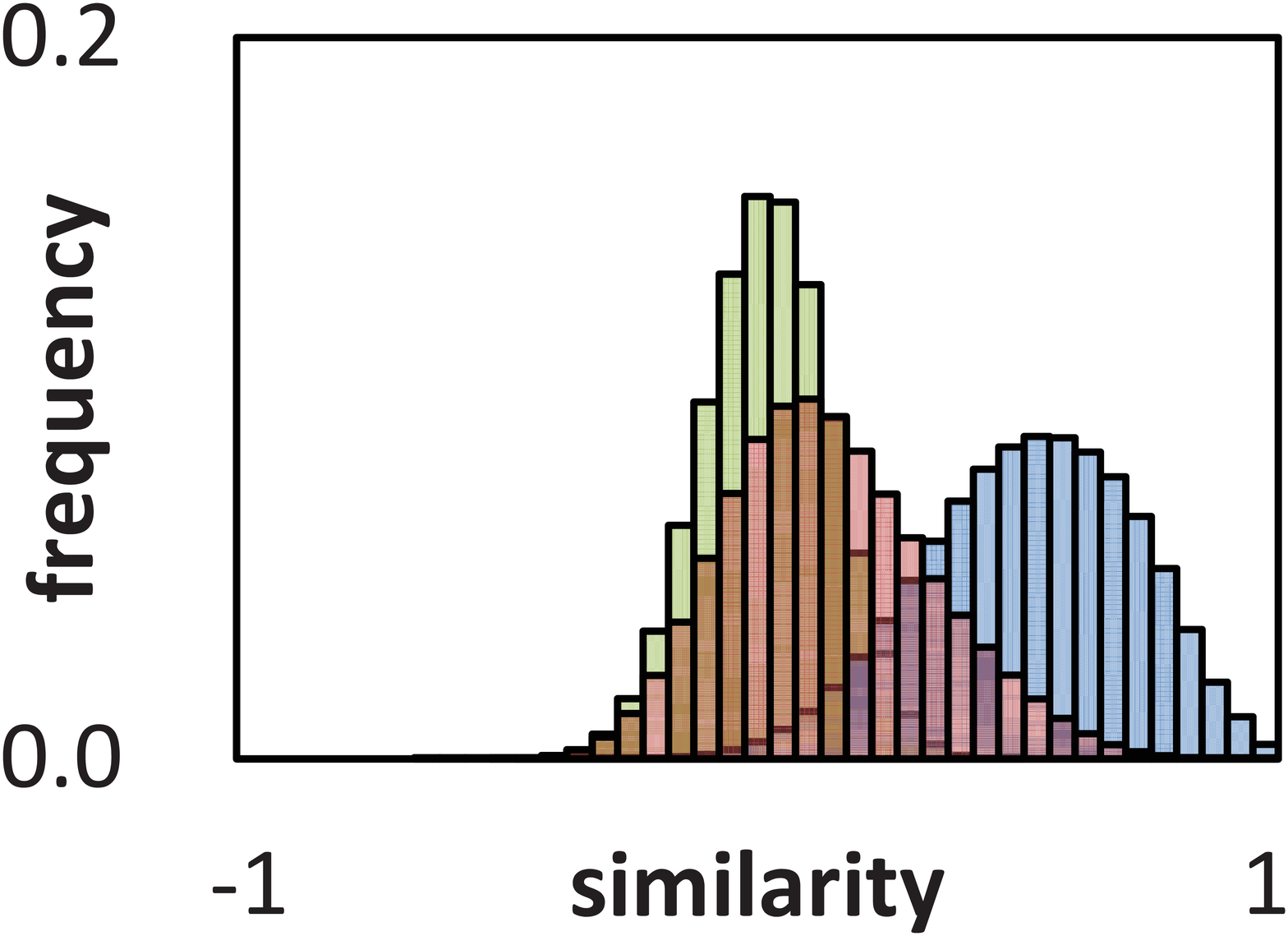}}
    }
\end{center}
\vspace{-7mm}
\caption{Histograms of cosine similarity of positive (blue), negative (red), self (green) pairs trained with different methods. Self pair refers to the pair of feature embeddings from different learners using same image. (a) Attention-based ensemble (ABE-$8$) using proposed loss, (b) attention-based ensemble (ABE-$8$) without divergence loss, (c) $8$-heads ensemble, (d) $8$-heads ensemble with divergence loss. In the case of attention-based ensemble, divergence loss is necessary for each learner to be trained to produce different features by attending to different locations. Without divergence loss, one can see all learners learn very similar embedding. Meanwhile, in the case of $M$-heads ensemble, there is no effect of applying divergence loss.
}
%\vspace{-4mm}
\label{fig:histogram}
\end{figure}

\subsubsection{ABE-$M$ without divergence loss}

\setlength{\tabcolsep}{4pt}
\begin{table}
\begin{center}
%\vspace{-5mm}
\caption{Recall@$K$(\%) comparison in ABE-$M$ ensemble without divergence loss $L_\mathrm{div}$ on CARS-196}
\vspace{-1mm}
\label{table:divlossabeeffect}
%\fontsize{9}{10}\selectfont
\scriptsize
\begin{tabular}{Sl  c c c c c c c c c}
\hline
& \multicolumn{4}{c}{{Ensemble}} && \multicolumn{4}{c}{{Individual Learners}}  \\
\cline{2-5}
\cline{7-10}
$K$ & 1 & 2 & 4 & 8 && 1 & 2 & 4 & 8\\
\hline
ABE-8$^{512}$  & 85.2 & 90.5 & 93.9 & 96.1 && 75.0\tiny{$\pm$0.39} & 83.4\tiny{$\pm$0.24} & 89.2\tiny{$\pm$0.31} & 93.2\tiny{$\pm$0.24}\\
\begin{tabular}{@{}l@{}} ABE-8$^{512}$\\ without $L_\mathrm{div}$ \end{tabular} & 69.7 & 78.8 & 86.2 & 91.5 && 69.5\tiny{$\pm$0.11} & 78.8\tiny{$\pm$0.14} & 86.1\tiny{$\pm$0.15} & 91.5\tiny{$\pm$0.09}\\
\hline
\end{tabular}
\end{center}
\vspace{-3mm}
\end{table}
\setlength{\tabcolsep}{1.4pt}

\begin{figure}[t]
\centering
\includegraphics[width=0.9\linewidth]{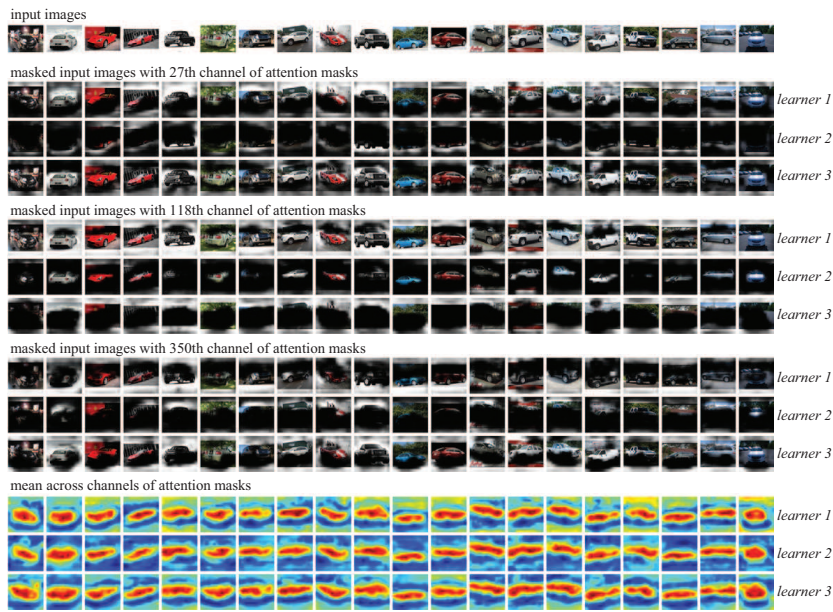}
%\vspace{-3mm}
\caption{The attention masks learned by each learner of ABE-8 on CARS-196. Due to the space limitation, results from only three learners out of eight and three channels out of 480 are illustrated. Each column shows the result of different input images. Different learners attend to different parts of the car such as upper part, bottom part, roof, tires, lights and so on}
%\vspace{-2mm}
\label{fig:mask}
\end{figure}

To analyze the effectiveness of divergence loss in ABE-$M$, we conduct experiments without divergence loss on CARS-196 and show the results in Table~\ref{table:divlossabeeffect}.
As we can see, ABE-$M$ without divergence loss performs similar to its individual learners whereas there is significant gain in ensemble performance of ABE-$M$ compared to its individual learners.

We also calculate the cosine similarity between positive, negative, and self pairs, and plot in Fig.~\ref{fig:histogram}.
With divergence loss (Fig.~\ref{fig:histogram_a}), all learners learn diverse embedding function which leads to decrease in cosine similarity of self pairs.
Without divergence loss (Fig.~\ref{fig:histogram_b}), all learners converge to very similar embedding function so that the cosine similarity of self pairs is close to 1.
This could be because all learners end up learning similar attention masks which leads to similar embeddings for all of them.

We visualize the learned attention masks of ABE-8 on CARS-196 in Fig.~\ref{fig:mask}.
Due to the space limitation, results from only three learners out of eight and three channels out of 480 are illustrated.
The figure shows that different learners are attending to different parts for the same channel.
Qualitatively, our proposed loss successfully diversify the attention masks produced by different learners.
They are attending to different parts of the car such as upper part, bottom part, roof, tires, lights and so on.
In 350th channel, for instance, learner 1 is focusing on bottom part of car, learner 2 on roof and learner 3 on upper part including roof.
At the bottom of Fig.~\ref{fig:mask}, the mean of the attention masks across all channels shows that the learned embedding function focuses more on object areas than the background.

\vspace{-6mm}
\subsubsection{Divergence loss in $M$-heads}
We show the result of experiments of 8-heads ensemble with divergence loss in Table~\ref{table:divlosseffect}.
We can see that the divergence loss does not improve the performance in 8-heads.
From Fig.~\ref{fig:histogram_c}, we can notice that cosine similarities of self pairs are close to zero for $M$-heads.
%This is because for $M$-heads, we train multiple global feature embedding functions $G_\ensemblesub(\cdot)$ for different learners.
Fig.~\ref{fig:histogram_d} shows that the divergence loss does not affect
 the cosine similarity of self pairs significantly.
As mentioned in Sec.~\ref{sec:ensemble_dml}, we hypothesize this is because each of $G_\ensemblesub (\cdot)$ could arbitrarily compose different metric spaces in $\mathcal{Y}$.

\setlength{\tabcolsep}{4pt}
\begin{table}
\begin{center}
%\vspace{-1mm}
\caption{Recall@$K$(\%) comparison in $M$-heads ensemble with divergence loss $L_\mathrm{div}$ on CARS-196}
\label{table:divlosseffect}
%\fontsize{9}{10}\selectfont
\scriptsize
\begin{tabular}{l  c c c c}
\hline
$K$ & 1 & 2 & 4 & 8 \\
\hline
8-heads & 76.1 & 84.3 & 90.3 & 93.9 \\
8-heads with $L_\mathrm{div}$ & 76.0 & 84.6 & 89.7 & 93.5 \\
\hline
\end{tabular}
\end{center}
%\vspace{-8mm}
\end{table}
\setlength{\tabcolsep}{1.4pt}

\vspace{-10mm}
\subsection{Ablation study} \label{sec:ablation}
%\vspace{-2mm}

To analyze the importance of various aspects of our model, we performed experiments on CARS-196 dataset of ABE-$8$ model, varying a few hyperparameters at a time and keeping others fixed. (More ablation study can be found in the supplementary material.)

\subsubsection{Sensitivity to depth of attention module}
\begin{figure}[t]
\begin{center}
    \subfigure[]{ \label{fig:numblocks} \includegraphics[width=0.3\linewidth]{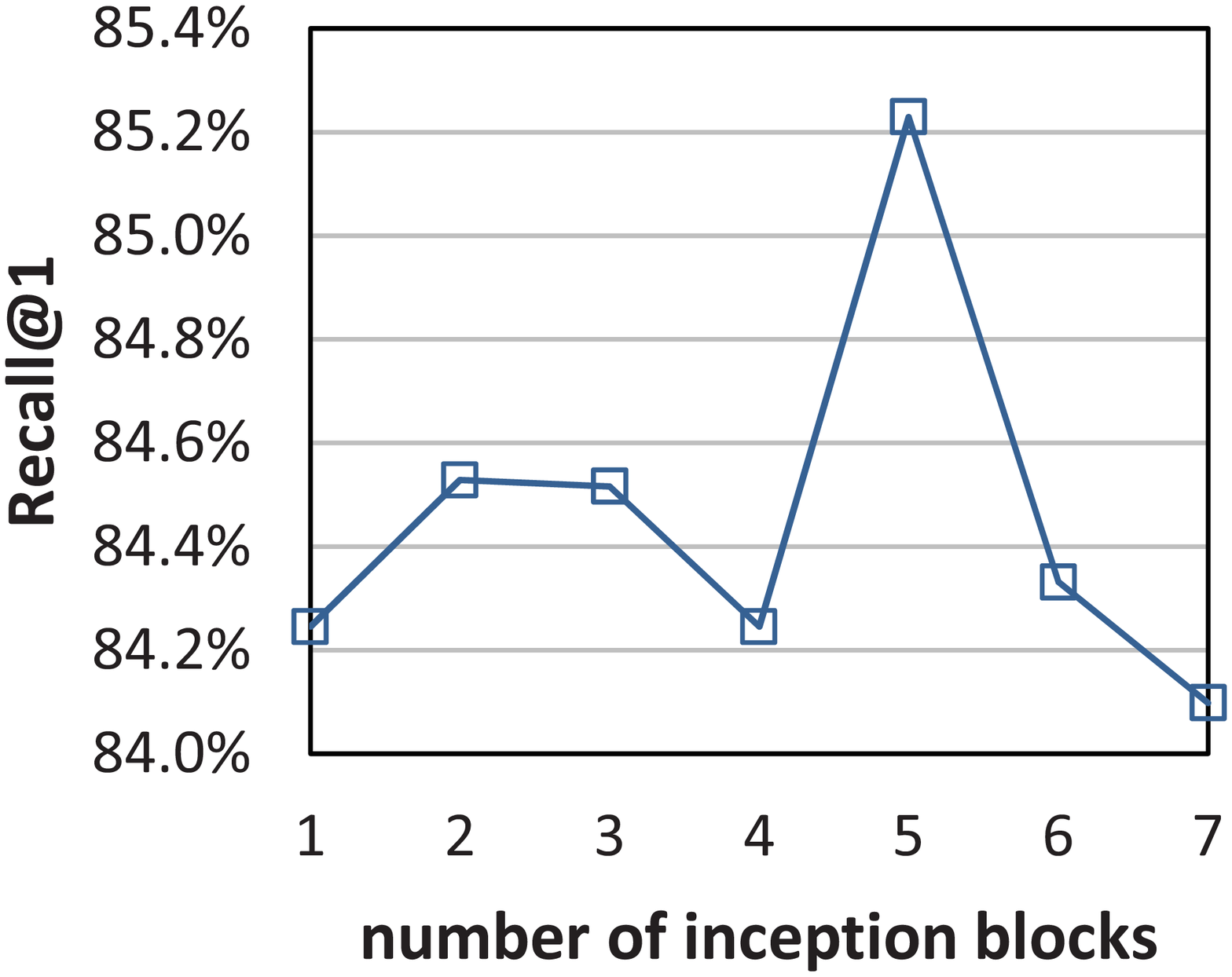}}
    \subfigure[]{ \label{fig:branchingpoint} \includegraphics[width=0.3\linewidth]{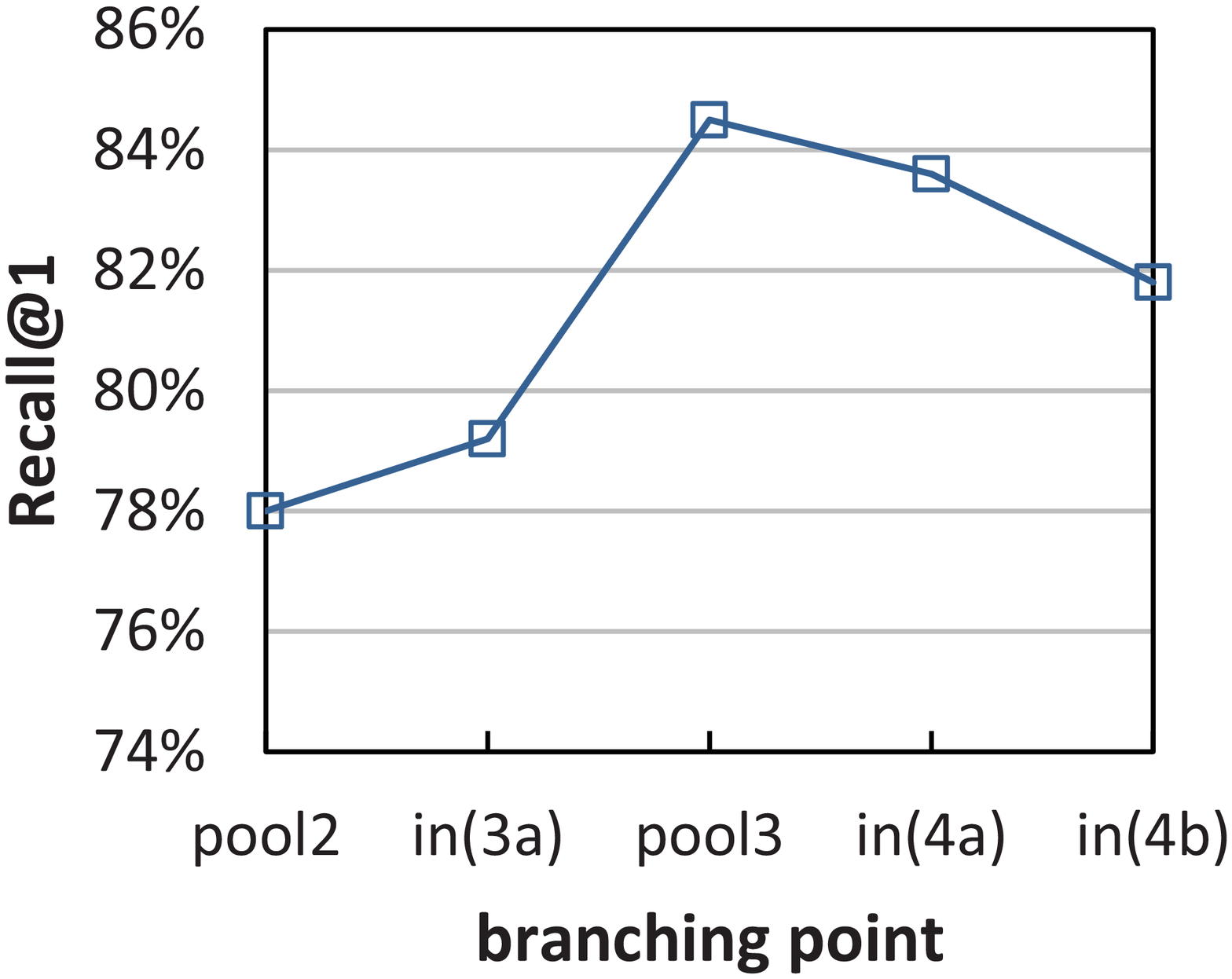}}
    \subfigure[]{ \label{fig:divweights} \includegraphics[width=0.3\linewidth]{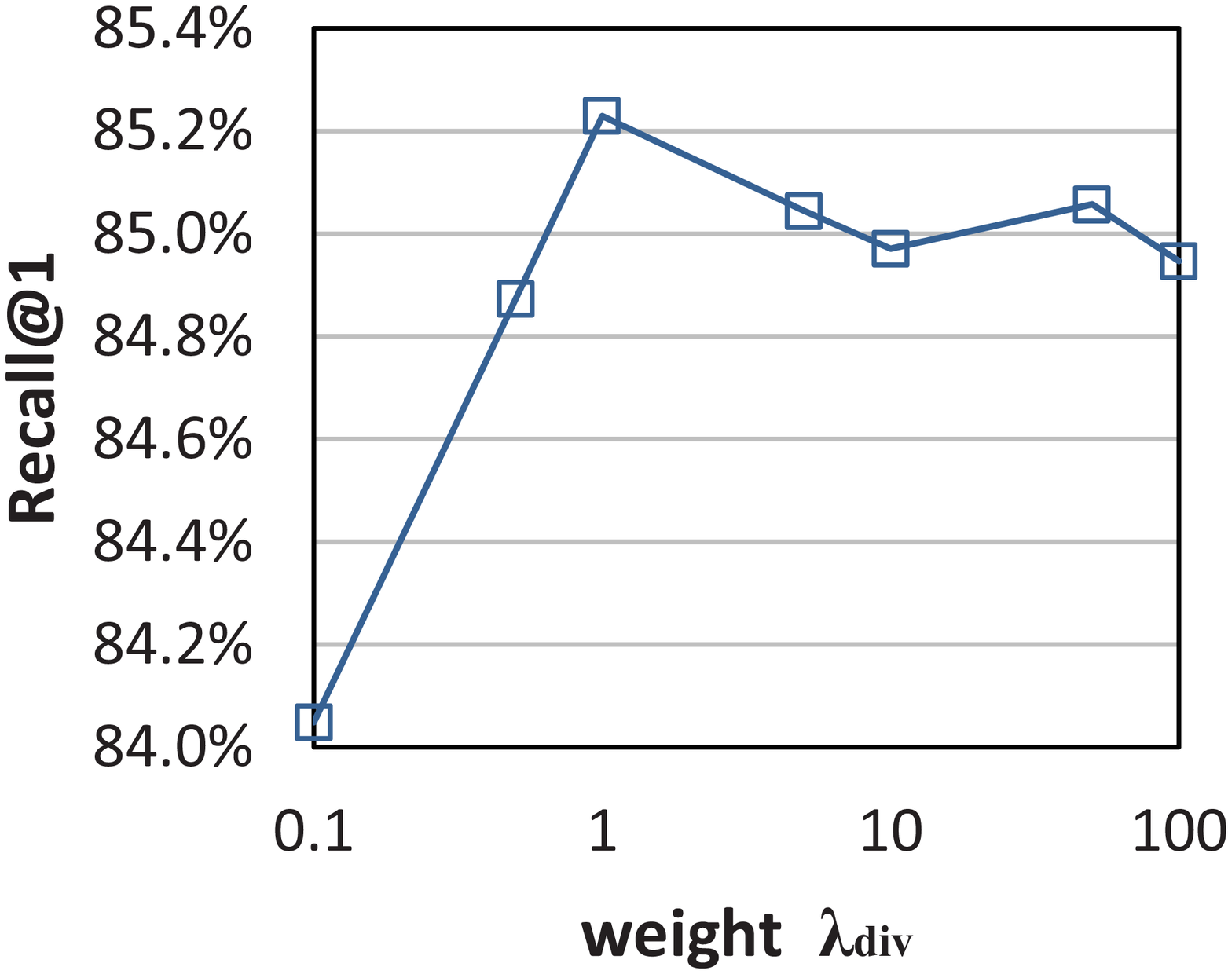}}
\end{center}
\vspace{-6mm}
\caption{Recall@1 while varying hyperparameters and architectures: (a) number of inception blocks used for attention module $A_k(\cdot)$, (b) branching point of attention module, and (c) weight $\lambda_\mathrm{div}$. Here, \texttt{inception(3a)} is abbreviated as \texttt{in(3a)}}
\label{fig:ablationfigures}
\end{figure}

We demonstrate the effect of depth of attention module by changing the number of inception blocks in it.
To make sure that we can take the element wise product of the attention mask with the input of attention module, the dimension of attention mask should match the input dimension of attention module.
Because of this we remove all the pooling layers in our attention module.
Fig.~\ref{fig:numblocks} shows Recall@1 with varying number of inception blocks in attention module starting from 1 (\texttt{inception(4a)}) to 7 (\texttt{inception(4a)} to \texttt{inception(5b)}) in GoogLeNet.
We can see that the attention module with 5 inception blocks (\texttt{inception(4a)} to \texttt{inception(4e)}) performs the best.

\subsubsection{Sensitivity to branching point of attention module}
The branching point of the attention module is where we split the network between spatial feature extractor $S(\cdot)$ and global feature embedding function $G(\cdot)$.
To analyze the choice of branching point of the attention module, we keep the number of inception blocks in attention module same ($\ie$ 5) and change branching points from \texttt{pool2} to \texttt{inception(4b)}. From Fig. \ref{fig:branchingpoint}, we see that \texttt{pool3} performs the best with our architecture.

We carry out this experiment with batch size 40 for all the branching points.
For ABE-$M$ model, the memory requirement for the $G(\cdot)$ is $M$ times compared to the individual learner. Since early branching point increases the depth of $G(\cdot)$ while decreasing the depth for $S(\cdot)$, it would consequently increase the memory requirement of the whole network.
Due to the memory constraints of GPU, we started the experiments from branching points  \texttt{pool2} and adjusted the batch size.

\subsubsection{Sensitivity to $\lambda_\mathrm{div}$}
Fig.~\ref{fig:divweights} shows the effect of $\lambda_\mathrm{div}$ on Recall@$K$ for ABE-$M$ model. We can see that $\lambda_\mathrm{div}=1$ performs the best and lower values degrades the performance quickly.

\subsection{Comparison with state of the art}
We compare the results of our approach with current state-of-the-art techniques.
Our model performs the best on all the major benchmarks for image retrieval.
Table~\ref{table:carscubcomptable}, \ref{table:sopcomptable} and \ref{table:inshopcomptable} compare the results with previous methods such as LiftedStruct~\cite{oh2016deep}, HDC~\cite{yuan2016hard}, Margin\footnote[2]{\label{note1}All compared methods use GoogLeNet architecture except Margin which uses ResNet-50~\cite{he2016deep}  and Proxy-NCA uses IncpeptionBN~\cite{ioffe2015batch}}~\cite{wu2017sampling}, BIER~\cite{opitz2017bier}, and A-BIER~\cite{opitz2017bier} on CARS-196~\cite{KrauseStarkDengFei-Fei_3DRR2013}, CUB-200-2011~\cite{WahCUB_200_2011}, SOP~\cite{oh2016deep}, and in-shop clothes retrieval~\cite{liu2016deepfashion} datasets.
Results on the cropped datasets are listed in Table~\ref{table:carscubcroppedcomptable}.

\setlength{\tabcolsep}{4pt}
\begin{table}
\begin{center}
\vspace{-2mm}
\caption{Recall@$K$(\%) score on CUB-200-2011 and CARS-196}
\vspace{-3mm}
\label{table:carscubcomptable}
%\fontsize{9}{10}\selectfont
\scriptsize
\begin{tabular}{Sl cccc c cccc}
\hline
 & \multicolumn{4}{c}{CUB-200-2011} && \multicolumn{4}{c}{CARS-196} \\
\cline{2-5}
\cline{7-10}
$K$ & 1 & 2 & 4 & 8 && 1 & 2 & 4 & 8 \\
\hline
Contrastive$^{128}$ \cite{oh2016deep} & 26.4 & 37.7 & 49.8 & 62.3 && 21.7 & 32.3 & 46.1 & 58.9 \\
LiftedStruct$^{128}$ \cite{oh2016deep} & 47.2 & 58.9 & 70.2 & 80.2 && 49.0 & 60.3 & 72.1 & 81.5 \\
N-Pairs$^{64}$ \cite{sohn2016improved} & 51.0 & 63.3 & 74.3 & 83.2& & 71.1 & 79.7 & 86.5 & 91.6 \\
Clustering$^{64}$ \cite{song2017deep} & 48.2 & 61.4 & 71.8 & 81.9 && 58.1 & 70.6 & 80.3 & 87.8 \\
Proxy NCA\ref{note1}$^{64}$ \cite{movshovitz2017no} & 49.2 & 61.9 & 67.9 & 72.4 && 73.2 & 82.4 & 86.4 & 87.8 \\
Smart Mining$^{64}$ \cite{ben2017smart} & 49.8 & 62.3 & 74.1 & 83.3 && 64.7 & 76.2 & 84.2 & 90.2 \\
Margin\ref{note1}$^{128}$ \cite{wu2017sampling} & \textbf{63.6} & \textbf{74.4} & \textbf{83.1} & \textbf{90.0} && 79.6 & 86.5 & 91.9 & 95.1 \\
HDC$^{384}$ \cite{yuan2016hard} & 53.6 & 65.7 & 77.0 & 85.6 && 73.7 & 83.2 & 89.5 & 93.8 \\
Angular Loss$^{512}$ \cite{wang2017deep} & 54.7 & 66.3 & 76.0 & 83.9 && 71.4 & 81.4 & 87.5 & 92.1 \\
A-Bier$^{512}$ \cite{opitz2018deep} & 57.5 & 68.7 & 78.3 & 86.2 && 82.0 & 89.0 & \underline{93.2} & \textbf{96.1} \\
\hline
ABE-2$^{384}$ & 55.9 & 68.1 & 77.4 & 85.7 && 77.2 & 85.1 & 90.5 & 94.2 \\
ABE-4$^{384}$ & 57.8 & 69.0 & 78.8 & 86.5 && 82.2 & 88.6 & 92.6 & 95.6 \\
\textbf{ABE-8$^{384}$} & 60.2 & 71.4 & \underline{80.5} & \underline{87.7} && \underline{83.8} & \underline{89.7} & \underline{93.2} & 95.5 \\
ABE-2$^{512}$ & 55.7 & 67.9 & 78.3 & 85.5 && 76.8 & 84.9 & 90.2 & 94.0 \\
ABE-4$^{512}$ & 57.9 & 69.3 & 79.5 & 86.9 && 82.5 & 89.1 & 93.0 & 95.5 \\
\textbf{ABE-8$^{512}$} & \underline{60.6} & \underline{71.5} & 79.8 & 87.4 && \textbf{85.2} & \textbf{90.5} & \textbf{94.0} & \textbf{96.1} \\
\hline
\end{tabular}
\vspace{-2mm}
\end{center}
\end{table}
\setlength{\tabcolsep}{1.4pt}

\setlength{\tabcolsep}{4pt}
\begin{table}
\begin{center}
\vspace{-2mm}
\caption{Recall@$K$(\%) score on CUB-200-2011 (cropped) and CARS-196 (cropped)}
\vspace{-3mm}
\label{table:carscubcroppedcomptable}
%\fontsize{9}{10}\selectfont
\scriptsize
\begin{tabular}{Sl cccc c cccc}
\hline
& \multicolumn{4}{c}{CUB-200-2011}&& \multicolumn{4}{c}{CARS-196} \\
\cline{2-5}
\cline{7-10}
$K$ & 1 & 2 & 4 & 8 && 1 & 2 & 4 & 8 \\
\hline
PDDM + Triplet$^{128}$ \cite{huang2016local} & 50.9 & 62.1 & 73.2 & 82.5 && 46.4&58.2&70.3&80.1 \\
PDDM + Quadruplet$^{128}$ \cite{huang2016local} & 58.3 & 69.2 & 79.0 & 88.4 && 57.4&68.6&80.1&89.4 \\
HDC$^{384}$ \cite{yuan2016hard} & 60.7 & 72.4 & 81.9 & 89.2 && 83.8&89.8&93.6&96.2 \\
Margin\ref{note1}$^{128}$ \cite{wu2017sampling} & 63.9 & 75.3 & 84.4 & 90.6 && 86.9&92.7&95.6&97.6 \\
A-BIER$^{512}$ \cite{opitz2018deep} & 65.5 & 75.8 & 83.9 & 90.2 && 90.3&94.1&\underline{96.8}&\underline{97.9} \\
ABE-2$^{512}$ & 64.9 & 76.2 & 84.2 & 90.0 && 88.2&92.8&95.6&97.3 \\
ABE-4$^{512}$ & \underline{68.0} & \underline{77.8} & \underline{86.3} & \underline{92.1} && \underline{91.6} & \underline{95.1} & \underline{96.8} &97.8 \\
\textbf{ABE-8$^{512}$} & \textbf{70.6} & \textbf{79.8} & \textbf{86.9} & \textbf{92.2} && \textbf{93.0} & \textbf{95.9} & \textbf{97.5} & \textbf{98.5} \\
\hline
\end{tabular}
\vspace{-2mm}
\end{center}
\end{table}
\setlength{\tabcolsep}{1.4pt}

\setlength{\tabcolsep}{4pt}
\begin{table}
\begin{center}
%\vspace{-2mm}
\caption{Recall@$K$(\%) score on Stanford online products dataset (SOP)}
\vspace{-1mm}
\label{table:sopcomptable}
%\fontsize{9}{10}\selectfont
\scriptsize
\begin{tabular}{Sl cccc}
\hline
$K$ & 1 & 10 & 100 & 1000 \\
\hline
Contrastive$^{128}$ \cite{oh2016deep} & 42.0 & 58.2 & 73.8 & 89.1 \\
LiftedStruct$^{512}$ \cite{oh2016deep} & 62.1 & 79.8 & 91.3 & 97.4 \\
N-Pairs$^{512}$ \cite{sohn2016improved} & 67.7 & 83.8 & 93.0 & 97.8 \\
Clustering$^{64}$ \cite{song2017deep} & 67.0 & 83.7 & 93.2 & - \\
Proxy NCA\ref{note1}$^{64}$ \cite{movshovitz2017no} & 73.7 & - & - & - \\
Margin\ref{note1}$^{128}$ \cite{wu2017sampling} & 72.7 & 86.2 & 93.8 & 98.0 \\
HDC$^{384}$ \cite{yuan2016hard} & 69.5 & 84.4 & 92.8 & 97.7 \\
A-Bier$^{512}$ \cite{opitz2018deep} & 74.2 & 86.9 & 94.0 & 97.8 \\
\hline
ABE-2$^{512}$ & 75.4 & 88.0 & 94.7 & \textbf{98.2} \\
ABE-4$^{512}$ & \underline{75.9} & \underline{88.3} & \underline{94.8} & \textbf{98.2} \\
\textbf{ABE-8$^{512}$} & \textbf{76.3} & \textbf{88.4} & \textbf{94.8} & \textbf{98.2} \\

\hline
\end{tabular}
%\vspace{-2mm}
\end{center}
\end{table}
\setlength{\tabcolsep}{1.4pt}

\setlength{\tabcolsep}{4pt}
\begin{table}
%\vspace{-2mm}
\begin{center}
\caption{Recall@$K$(\%) score on in-shop clothes retrieval dataset}
\vspace{-1mm}
\label{table:inshopcomptable}
%\fontsize{8}{10}\selectfont
\scriptsize
\begin{tabular}{Sl cccccc}
\hline
$K$ & 1 & 10 & 20 & 30 & 40 & 50\\
\hline
FasionNet+Joints$^{4096}$ \cite{liu2016deepfashion} & 41.0 & 64.0 & 68.0 & 71.0 & 73.0 & 73.5  \\
FasionNet+Poselets$^{4096}$ \cite{liu2016deepfashion} & 42.0 & 65.0 & 70.0 & 72.0 & 72.0 & 75.0  \\
FasionNet$^{4096}$ \cite{liu2016deepfashion} & 53.0 & 73.0 & 76.0 & 77.0 & 79.0 & 80.0  \\
HDC$^{384}$ \cite{yuan2016hard} & 62.1 & 84.9 & 89.0 & 91.2 & 92.3 & 93.1  \\
A-BIER$^{512}$ \cite{opitz2018deep} & 83.1 & 95.1 & 96.9 & 97.5 & 97.8 & 98.0  \\
\hline
ABE-2$^{512}$ & 85.2 & 96.0 & 97.2 & 97.8 & 98.2 & 98.4  \\
ABE-4$^{512}$ & \underline{86.7} & \underline{96.4} & \underline{97.6} & \underline{98.0} & \underline{98.4} & \underline{98.6}  \\
\textbf{ABE-8$^{512}$} & \textbf{87.3} & \textbf{96.7} & \textbf{97.9} & \textbf{98.2} & \textbf{98.5} & \textbf{98.7}  \\

\hline
\end{tabular}
\vspace{-2mm}
\end{center}
\end{table}
\setlength{\tabcolsep}{1.4pt}

%% file: conclusion.tex
\section{Conclusion}
In this work, we present a new framework for ensemble in the domain of deep metric learning.
It uses attention-based architecture that attends to parts of the image.
We use multiple such attention-based learners for our ensemble.
Since ensemble benefits from diverse learners,
we further introduce a divergence loss to diversify the feature embeddings learned by each learner.
The divergence loss encourages that the attended parts of the image for each learner are different.
Experimental results demonstrate that the divergence loss not only increases the performance of ensemble
but also increases each individual learners' performance compared to the baseline.
We demonstrate that our method outperforms the current state-of-the-art techniques by significant margin
on several image retrieval benchmarks including CARS-196 \cite{KrauseStarkDengFei-Fei_3DRR2013},
CUB-200-2011 \cite{WahCUB_200_2011}, SOP \cite{oh2016deep}, and in-shop clothes retrieval \cite{liu2016deepfashion} datasets.

%% file: appendix.tex
%\section{Appendix}

\title{ \large{Attention-based Ensemble for Deep Metric Learning}\\ \large{(Supplementary Material)}}

\titlerunning{Attention-based Ensemble for Deep Metric Learning}

\authorrunning{W. Kim, B. Goyal, K. Chawla, J. Lee, K. Kwon}

\author{Wonsik Kim, Bhavya Goyal, Kunal Chawla, Jungmin Lee, Keunjoo Kwon}
\institute{Samsung Research, \\
Samsung Electronics\\
\email{ \{wonsik16.kim, bhavya.goyal, kunal.chawla, jm411.lee, keunjoo.kwon\}@samsung.com}
}

\maketitle

\vspace{-7mm}
\begin{figure}[h]
\begin{center}
    \mbox{%
    \subfigure[$3$-heads+att]{ \label{fig:gmsams} \includegraphics[height=0.29\linewidth]{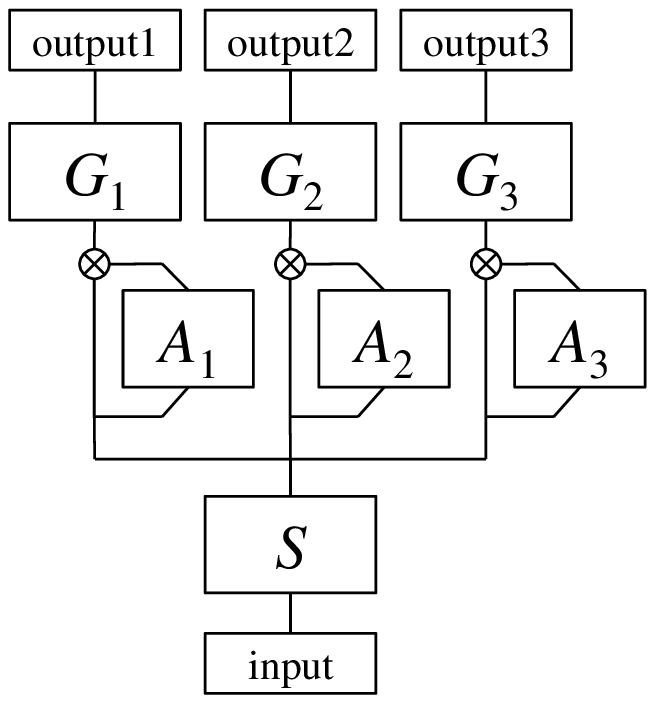}}
%    }
%    \mbox{%
\hspace{3mm}
    \subfigure[$3$-tails]{ \label{fig:gsm} \includegraphics[height=0.29\linewidth]{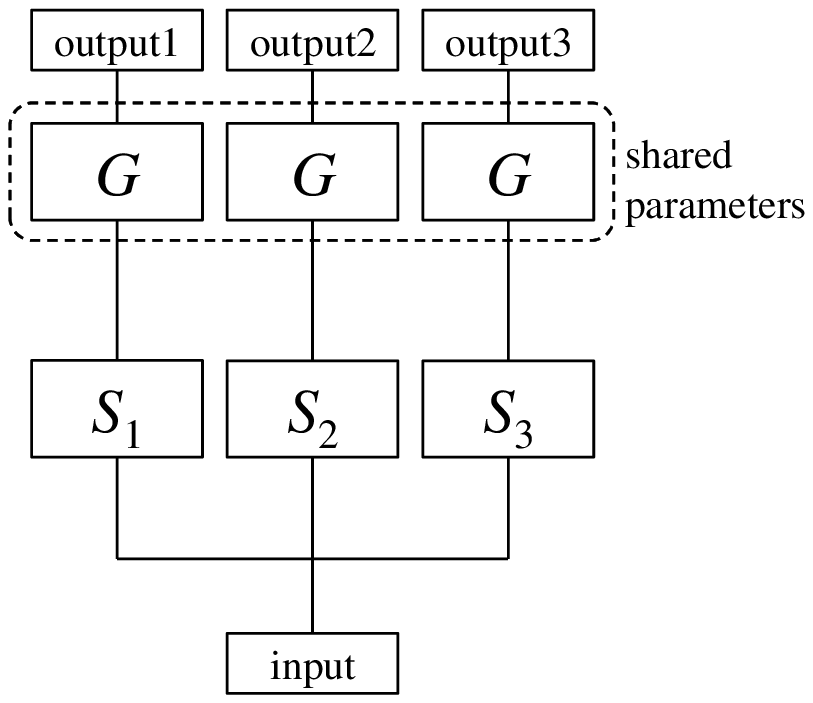}}
    }
\end{center}
\vspace{-7mm}
\caption{Illustration of additional architectures reported in this supplementary material. \text{(a) $3$-heads+att} ($3$-heads with attention module) and (b) $3$-tails (ensemble of $G(S_m(x))$)
}
\label{fig:m_heads_and_m_way}
\end{figure}

\noindent
In this supplementary material, we consider two additional architecutres for the ablation study: $M$-heads+att and $M$-tails. The details of each architectures are following. All experiments in this supplementary material are done on CARS-196 dataset. The models are trained with ensemble embedding size 512 and batch size 64, unless otherwise specified.

As described in Sec. 3 of the main manuscript, $S(\cdot)$ is the spatial feature extractor, $G(\cdot)$ is the global feature embedding function, and $A(\cdot)$ is the attention module.
Our combined embedding function $\Ensemblefunc_\ensemblesub(x)$ for the learner $\ensemblesub$ in ABE-$M$ is defined as the following:
%\vspace{-2mm}
\begin{equation}
	\Ensemblefunc_\ensemblesub(x) = G(S(x) \circ A_\ensemblesub(S(x)) ),
\label{eq:embedding}
%\vspace{-2mm}
\end{equation}
where $\circ$ denotes element-wise product.

\begin{itemize}
%\item \textbf{$M$-models}  It refers to the ensemble of $F_\ensemblesub(x)$ = $G_\ensemblesub(S_\ensemblesub(x))$. We compare our results with naive ensembling technique where all learners are trained independently. It is trained without divergence loss.

\item \textbf{$M$-heads+att} It refers to the $M$-heads with attention model, which
 is the ensemble of $G_\ensemblesub(S(x) \circ A_\ensemblesub(S(x))$ (Fig. \ref{fig:gmsams}).
Here, unlike our ABE-$M$ model, the global feature embedding function
is not shared and the model is trained without divergence loss.

\item \textbf{$M$-tails} Instead of having multiple heads and a shared tail, we can
consider multiple tails and a shared head, which is $G(S_\ensemblesub(x))$ (Fig. \ref{fig:gsm}).
Unlike ABE-$M$, there are no attention modules, and the spatial
feature extractor is not shared, but the global feature embedding
function is. It is also trained with divergence loss. Due to the memory constraint,
this model is trained with a batch size of 32.

\end{itemize}

\setlength{\tabcolsep}{4pt}
\begin{table}
\begin{center}
%\vspace{-2mm}
\caption{Recall@$K$(\%) comparison on CARS-196. All presented methods use ensemble embedding size of 512}
%\vspace{-3mm}
\label{table:baselinecomptable}
%\fontsize{9}{10}\selectfont
\scriptsize
\begin{tabular}{Sl   c c c c      c     c c c c    c c }
\hline
 & \multicolumn{4}{c}{{Ensemble}} && \multicolumn{4}{c}{{Individual Learners}} & \tiny{params} &  \tiny{flops} \\
\cline{2-5}
\cline{7-10}
\hspace{-1mm}$K$ & 1 & 2 & 4 & 8 & & 1 & 2 & 4 & 8  & \tiny{($\times 10^7$)}  &  \tiny{($\times 10^9$)}   \\
\hline
\hspace{-1mm}8-heads  & 76.1 & 84.3 & 90.3 & 93.9 && 68.3\tiny{$\pm$.39} & 78.5\tiny{$\pm$.39} &86.0\tiny{$\pm$.37} &91.3\tiny{$\pm$.31} & 4.36 & 6.28\\
\hspace{-1mm}8-heads+att  & 79.7 & 87.0 & 91.6 & 94.8 && 69.0\tiny{$\pm$.36} & 78.9\tiny{$\pm$.27} &86.2\tiny{$\pm$.30} &91.4\tiny{$\pm$.21} & 4.96 & 7.46\\
\hspace{-1mm}8-tails  & 81.1 & 88.0 & 92.4 & 95.4 && 71.4\tiny{$\pm$.83} & 80.9\tiny{$\pm$.55} &87.5\tiny{$\pm$.31} &92.3\tiny{$\pm$.26} & 1.08 & 12.7\\
\hspace{-1mm}ABE-8  & 85.2 & 90.5 & 93.9 & 96.1 && 75.0\tiny{$\pm$.39} & 83.4\tiny{$\pm$.24} &89.2\tiny{$\pm$.31} &93.2\tiny{$\pm$.24} & 1.20 & 7.46\\

\hline
\end{tabular}
\end{center}
\end{table}
\label{table:baseline}
%\vspace{-6mm}
\setlength{\tabcolsep}{1.4pt}

Table~\ref{table:baseline} summarizes the results. In addition, it also presents the results of $8$-heads ($G_\ensemblesub(S(x))$) and ABE-$8$ for comparison.

\subsubsection{Effect of attention module on $M$-heads}
\label{sec:m-heads-att}

To study the contribution of attention module on performance, we propose
$M$-heads with attention model ($M$-heads+att). This model has multiple heads $G_m(\cdot)$
and shared spatial feature extractor $S(\cdot)$, like $M$-heads model. In addition we attach $M$
different attention models for $M$ different learners, similar to our proposed ABE-$M$.
There are two differences between $M$-heads+att and ABE-$M$. Firstly, the global
 embedding function $G(\cdot)$ is not shared. Second, divergence loss is not applied during
training. As discussed in the main manuscript, divergence loss is not required
when the global embedding function $G_m(\cdot)$ is different for each learner.

The results show that with attention module, Recall@1 is improved from
76.1$\%$ to 79.7$\%$ compared to $8$-heads. This comparison shows the effect of
attention module independent from other factors. Along with the attention module,
 by sharing the global embedding function $G(\cdot)$ and applying divergence loss, ABE-$8$
 further improves Recall@1 to 85.2$\%$.

\subsubsection{Comparison of $M$-heads and $M$-tails}
\label{sec:m-heads-tails}

The main manuscript considers two different ways of ensembles in regards
 to two-step embedding function. In two-step embedding function, input is first mapped
 to an intermediate metric space by $S(\cdot)$ and then mapped to the final embedding
space by $G(\cdot)$. The first way of ensemble shares $S(\cdot)$ while having different $G_m(\cdot)$
 for each learner. The second way shares $G(\cdot)$ while having
different $S_m(\cdot)$ for each learner. To investigate the effect of two different ways of
 ensembles on performance, we propose $M$-tails model as depicted in Fig. \ref{fig:gmsams}.

The results demonstrate $8$-tails achieves better Recall@1 compared to $8$-heads.
The performance gain from this architecture (+5.0$\%$) is even larger
than the case of $8$-heads+att (+3.6$\%$). Compared to $8$-tails, ABE-$8$
further improves the performance by using attention module.